\documentclass[twoside,11pt]{article}

\usepackage[nohyperref, preprint]{jmlr2e}

\usepackage[utf8]{inputenc} 
\usepackage[T1]{fontenc}    
\usepackage[english]{babel} 
\usepackage{microtype}      
\usepackage{textgreek}      
\usepackage[verbose]{newunicodechar} 
\usepackage{xspace}
\usepackage{textcomp}       

\usepackage{subfigure}
\usepackage[rightcaption]{sidecap}  
\usepackage{booktabs}       
\usepackage{caption}     

\sidecaptionvpos{figure}{t} 

\usepackage{rotating}

\usepackage{amsfonts}       
\usepackage{amsmath}        
\usepackage{nicefrac}       
\usepackage[table]{xcolor}	
\usepackage{bm}             

\definecolor{mydarkblue}{rgb}{0,0.08,0.45}

\definecolor{viridis_petrol}{rgb}{0.12,0.5,0.47}
\definecolor{viridis_blue}{rgb}{0.15,0.33,0.48}
\definecolor{viridis_green}{rgb}{0.5,0.82,0.21}

\definecolor{gs_purple}{rgb}{0.30,0.28,0.64}
\definecolor{gs_red}{rgb}{0.84,0.29,0.14}
\definecolor{gs_green}{rgb}{0.31,0.51,0.47}

\usepackage{url}  

\usepackage{varioref}
\usepackage{hyperref}

\hypersetup{ %
    breaklinks=true,
    unicode=true,
    pdftitle={Binding Survey},
    pdfauthor={Klaus Greff},
    pdfsubject={},
    pdfkeywords={Binding},
    pdfborder=0 0 0,
    pdfpagemode=UseNone,
    colorlinks=true,
    linkcolor=mydarkblue,
    citecolor=mydarkblue,
    filecolor=mydarkblue,
    urlcolor=mydarkblue,
    pdfview=FitH}
\usepackage[all]{hypcap}
\usepackage[nameinlink]{cleveref}

\crefname{chapter}{Chapter}{Chapters}
\crefname{section}{Section}{Sections}
\crefname{subsection}{Section}{Sections}
\crefname{subsubsection}{Section}{Sections}
\crefname{figure}{Figure}{Figures}
\crefname{table}{Table}{Tables}
\crefname{subfigure}{Figure}{Figures}
\crefname{page}{Page}{Pages}
\crefname{equation}{Equation}{Equations}
\crefname{appendix}{Appendix}{Appendix}

\crefformat{footnote}{#2\footnotemark[#1]#3}

\newunicodechar{ℕ}{\ensuremath{\mathbb{N}}}
\newunicodechar{ℤ}{\ensuremath{\mathbb{Z}}}
\newunicodechar{ℝ}{\ensuremath{\mathbb{R}}}
\newunicodechar{ℂ}{\ensuremath{\mathbb{C}}}

\newunicodechar{⊂}{\subset}
\newunicodechar{∈}{\in}
\newunicodechar{∪}{\cup}
\newunicodechar{∩}{\cap}

\newunicodechar{∀}{\forall}
\newunicodechar{∃}{\exists}
\newunicodechar{¬}{\neg}
\newunicodechar{∨}{\wedge}
\newunicodechar{∧}{\vee}

\newcommand{\textOrMath}[2]{{\ifmmode{#1}\else{#2}\fi}}

\newunicodechar{⇐}{\Leftarrow}
\newunicodechar{⇔}{\Leftrightarrow}
\newunicodechar{⇒}{\Rightarrow}
\newunicodechar{→}{\rightarrow}
\newunicodechar{↦}{\mapsto}

\newunicodechar{√}{\srqt}
\newunicodechar{ℵ}{\aleph}
\newunicodechar{×}{\times}
\newunicodechar{∂}{\partial}
\newunicodechar{∘}{\circ}
\newunicodechar{∇}{\nabla}
\newunicodechar{⊥}{\perp}
\newunicodechar{∡}{\measuredangle}
\newunicodechar{∥}{\|}
\newunicodechar{∞}{\infty}
\newunicodechar{∝}{\propto}
\newunicodechar{∅}{\varnothing}

\makeatletter
\newcommand*{\etc}{%
    \protect\@ifnextchar{.}%
        {etc}%
        {etc.\@\xspace}%
}
\newcommand*{\eg}{%
    \protect\@ifnextchar{.}%
        {e.g}%
        {e.g.\@\xspace}%
}
\newcommand*{\ie}{%
    \protect\@ifnextchar{.}%
        {i.e}%
        {i.e.\@\xspace}%
}
\newcommand*{\cf}{%
    \protect\@ifnextchar{.}%
        {cf}%
        {cf.\@\xspace}%
}
\newcommand*{\wrt}{%
    \protect\@ifnextchar{.}%
        {wrt}%
        {wrt.\@\xspace}%
}
\makeatother

\jmlrheading{21}{2020}{??-??}{??/??}{??/??}{paperid}{Klaus Greff and Sjoerd van Steenkiste and Jürgen Schmidhuber}

\ShortHeadings{On the Binding Problem in Artificial Neural Networks}{Greff and van Steenkiste and Schmidhuber}

\newcommand\blfootnote[1]{%
  \begingroup
  \renewcommand\thefootnote{}\footnote{#1}%
  \addtocounter{footnote}{-1}%
  \endgroup
}

\makeatletter
\long\def\@makefntext#1{\@setpar{\@@par\@tempdima \hsize 
             \advance\@tempdima-15pt\parshape \@ne 15pt \@tempdima}\par
             \parindent 2em\noindent \hbox to \z@{\hss{\@thefnmark} \hfil}#1}
\makeatother

\begin{document}


\title{On the Binding Problem in Artificial Neural Networks}
\date{December 2020}
\author{\name Klaus Greff$^*$ \email klausg@google.com\\
		\addr Google Research, Brain Team\\
		Tucholskystraße 2, 10116 Berlin, Germany
        \AND \name Sjoerd van Steenkiste \email sjoerd@idsia.ch
        \AND \name Jürgen Schmidhuber \email juergen@idsia.ch\\
        \addr Istituto Dalle Molle di studi sull'intelligenza artificiale (IDSIA)\\
        Università della Svizzera Italiana (USI)\\
        Scuola universitaria professionale della Svizzera italiana (SUPSI)\\
        Via la Santa 1, 6962 Viganello, Switzerland}

\editor{Kevin Murphy and Bernhard Schölkopf}

\maketitle


\begin{abstract}
Contemporary neural networks still fall short of human-level generalization, which extends far beyond our direct experiences.
In this paper, we argue that the underlying cause for this shortcoming is their inability to dynamically and flexibly bind information that is distributed throughout the network.
This \emph{binding problem} affects their capacity to acquire a compositional understanding of the world in terms of symbol-like entities (like objects), which is crucial for generalizing in predictable and systematic ways.
To address this issue, we propose a unifying framework that revolves around forming meaningful entities from unstructured sensory inputs (segregation), maintaining this separation of information at a representational level (representation), and using these entities to construct new inferences, predictions, and behaviors (composition).
Our analysis draws inspiration from a wealth of research in neuroscience and cognitive psychology, and surveys relevant mechanisms from the machine learning literature, to help identify a combination of inductive biases that allow symbolic information processing to emerge naturally in neural networks.
We believe that a compositional approach to AI, in terms of grounded symbol-like representations, is of fundamental importance for realizing human-level generalization, and we hope that this paper may contribute towards that goal as a reference and inspiration.
\end{abstract}

\blfootnote{* This research was partially conducted while the author was affiliated with IDSIA, USI \& SUPSI.}\blfootnote{** A preliminary version of this work was presented at an ICML Workshop~\citep{vansteenkiste2019perspective}.}

\makeatletter
\long\def\@makefntext#1{\@setpar{\@@par\@tempdima \hsize 
             \advance\@tempdima-15pt\parshape \@ne 15pt \@tempdima}\par
             \parindent 2em\noindent \hbox to \z@{\hss{\@thefnmark}. \hfil}#1}
\makeatother


{

\section{Introduction}
\label{sec:intro}
Existing neural networks fall short of human-level generalization.
They require large amounts of data, struggle with transfer to novel tasks, and are fragile under distributional shift.
However, under the right conditions, they have shown a remarkable capacity for learning and modeling complex statistical structure in real-world data. 
One explanation for this discrepancy is that neural networks mostly learn about surface statistics in place of the underlying concepts, which prevents them from generalizing systematically.
However, despite considerable effort to address this issue, human-level generalization remains a major open problem.\looseness=-1


In this paper, we will view the inability of contemporary neural networks to effectively form, represent, and relate symbol-like entities, as the root cause of this problem.
This emphasis on symbolic reasoning reflects a common sentiment within the community and others have advocated similar perspectives~\citep{fodor1988connectionism,marcus2003algebraic,lake2017building}.
Indeed, it is well established that human perception is structured around objects, which serve as compositional `building blocks' for many aspects of higher-level cognition such as language, planning, and reasoning.
This understanding of the world, in terms of parts that can be processed independently and recombined in near-infinite ways, allows humans to generalize far beyond their direct experiences.


Meanwhile, the persistent failure of neural networks to generalize systematically is evidence that neural networks do not acquire the ability to process information symbolically, simply as a byproduct of learning.
Specialized inductive biases that mirror aspects of human information processing, such as attention or memory, have led to encouraging results in certain domains.
However, the general issue remains unresolved, which has led some to believe that the way forward is to build hybrid systems that combine connectionist methods with inherently symbolic approaches.
In contrast, we believe that these problems stem from a deeper underlying cause that is best addressed directly from \emph{within} the framework of connectionism.\looseness=-1


In this work, we argue that this underlying cause is the \emph{binding problem}: The inability of existing neural networks to dynamically and flexibly \emph{bind} information that is distributed throughout the network.
The binding problem affects their ability to form meaningful entities from unstructured sensory inputs (segregation), to maintain this separation of information at a representational level (representation), and to use these entities to construct new inferences, predictions, and behaviors (composition).
Each of these aspects relates to a wealth of research in neuroscience and cognitive psychology, where the binding problem has been extensively studied in the context of the human brain.
Based on these connections, we work towards a solution to the binding problem in neural networks and identify several important challenges and requirements.
We also survey relevant mechanisms from the machine learning literature that either directly or indirectly already address some of these challenges.
Our analysis provides a starting point for identifying the right combination of inductive biases to enable neural networks to process information symbolically and generalize more systematically.


In our view, integrating symbolic processing into neural networks is of fundamental importance for realizing human-level AI, and will require a joint community effort to resolve.
The goal of this survey is to support this effort, by organizing various related research into a unifying framework based on the binding problem. 
We hope that it may serve as an inspiration and reference for future work that bridges related fields and sparks fruitful discussions.\looseness=-1 

}

{
\section{The Binding Problem}
\label{sec:bind}

We start our discussion by reviewing the importance of symbols as units of computation and highlight several symptoms that point to the lack of emergent symbolic processing in existing neural networks. 
We argue that this is a major obstacle for achieving human-level generalization, and posit that the binding problem in connectionism is the underlying cause for this weakness.
This section serves as an introduction to the binding problem and provides the necessary context for the subsequent in-depth discussion of its individual aspects in \cref{sec:rep,sec:seg,sec:comp}.

\subsection{Importance of Symbols}
\label{sec:bind:symbols}

The human capacity to comprehend reaches far beyond direct experiences. 
We are able to reason causally about unfamiliar scenes, understand novel sentences with ease, and use models and analogies to make predictions about entities far outside the scope of everyday reality, like atoms, and galaxies.
This seemingly infinite expressiveness and flexibility of human cognition has long fascinated philosophers, psychologists, and AI researchers alike.
The best explanation for this remarkable cognitive capacity revolves around symbolic thought: the ability to form, manipulate, and relate mental entities that can be processed like symbols~\citep{whitehead1985symbolism}.
By decomposing the world in terms of abstract and reusable `building blocks’, humans are able to understand novel contexts in terms of known concepts, and thereby leverage their existing knowledge in near-infinite ways.
This compositionality underlies many high-level cognitive abilities such as language, causal reasoning, mathematics, planning, analogical thinking, \etc.

Human understanding of the world in terms of objects develops at an early age~\citep{spelke2007core} and infants as young as five months appear to understand that objects continue to exist in the absence of visual stimuli (object permanence; \citealp{baillargeon1985object}).
Arguably, this decoupling of mental representation from direct perception is a first step towards a compositional description of the world in terms of more abstract entities.
By the age of eighteen months, young children have acquired the ability to use gestures symbolically to refer to objects or events~\citep{acredolo1988symbolic}.
This ability to relate sensory entities is then key to the subsequent grounding of language. 
As the child grows up, entities become increasingly more general and start to include categories, concepts, events, behaviors, and other abstractions, together with a growing number of universal relations such as ``same'', ``greater than'', ``causes'', \etc.
This growing set of composable building blocks yields an increasingly more powerful toolkit for constructing structured mental models of the world~\citep{johnson-laird2010mental}.

The underlying compositionality of such symbols is equally potent for AI, and numerous methods that model intelligence as a symbol manipulation process have been explored.
Early examples included tree-search over abstract state spaces such as the General Problem Solver~\citep{newell1959report} for theorem proving, or chess~\citep{campbell2002deep}; Expert systems that made use of decision trees to perform narrow problem solving for hardware design~\citep{sollow1987assessing} and medical diagnosis~\citep{shortliffe1975computerbased}; Natural language parsers that used a dictionary and a fixed set of grammatical rules to interpret written English; And knowledge bases such as semantic networks (networks of concepts and relations) that could be used to answer basic questions~\citep{weizenbaum1966eliza}, solve basic algebra word problems~\citep{bobrow1964natural}, or control simple virtual block worlds~\citep{winograd1971procedures}.
All of these examples of \emph{symbolic AI} relied on manually designed symbols and rules of manipulation, which allowed them to generalize in \emph{predictable} and \emph{systematic} ways. 
Since then, many of these approaches have become part of the standard computer-science toolbox\footnote{They are hardly called AI anymore since it is now well understood how to solve the problems that they address. This redefinition of what constitutes AI is sometimes called the \emph{AI effect}, summarized by Douglas Hofstadter as ``AI is whatever hasn't been done yet''.}.

\begin{figure}
    \centering
    \includegraphics[width=\textwidth]{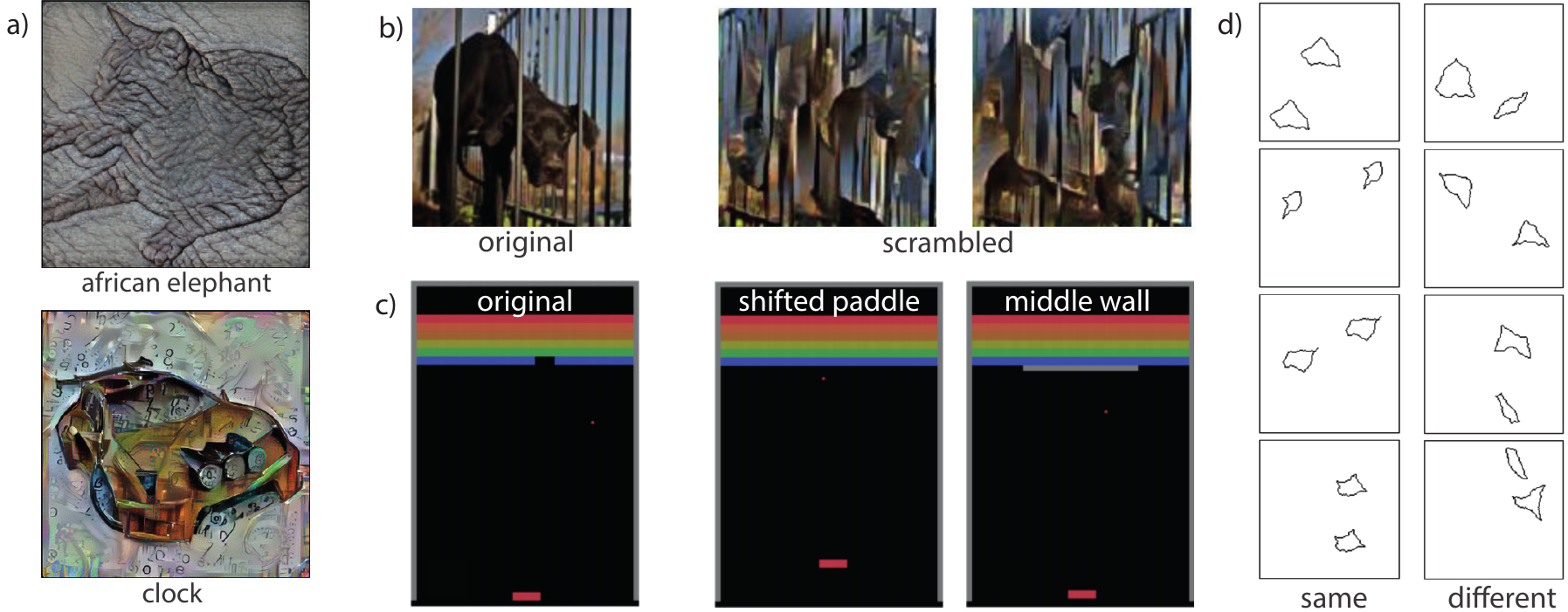}
    \caption{
        Various evidence for shortcomings of current neural networks.
        \textbf{(a)} CNN image classifiers are biased towards texture over shape~\citep{geirhos2018imagenettrained} and \textbf{(b)} can be well approximated by bag-of-local-features models~\citep{brendel2019approximating}. Hence, scrambling the image in a way that preserves local (but not global) structures affects them less than humans.
        \textbf{(c)} Neural network based agents trained on Breakout, fail to generalize to slight variations of the game such as a shifted paddle or an added middle wall~\citep{kansky2017schema}.
        \textbf{(d)} Neural networks also struggle to learn visual relations such as whether two shapes are the same or different~\citep{fleuret2011comparing,ricci2018notsoclevr}.
    }
    \label{fig:bind:failures}
\end{figure}

\subsection{Symbolic processing in Connectionist Methods}
\label{sec:bind:connectionism}

Connectionism takes a different, brain-inspired, approach to Artificial Intelligence that stands in contrast to symbolic AI and its focus on the conscious mind~\citep{newell1981computer,fodor1975language}.
Rather than relying on hand-crafted symbols and rules, connectionist approaches such as neural networks focus on \emph{learning} suitable distributed representations directly from low-level sensory data.
In this way, neural networks have resolved many of the problems that haunted symbolic AI, including their brittleness when confronted with inconsistencies or noise, and the prohibitive amount of human engineering and interpretation that would be required to apply these techniques on low-level perceptual tasks.
Importantly, the distributed representations learned by neural networks are directly grounded in their input data, unlike symbols whose connection to real-world concepts is entirely subject to human interpretation (see \emph{symbol grounding problem}; \citealp{harnad1990symbol}).
Modern neural networks have proven highly successful and superior to symbolic approaches in perceptual domains, such as in visual object recognition~\citep{ciresan2011flexible,ciresan2012multicolumn,krizhevsky2012imagenet} or speech recognition~\citep{fernandez2007sequence,hinton2012deep}, and even in some inherently symbolic domains such as language modeling~\citep{devlin2018bert,radford2019language,brown2020language}, translation~\citep{wu2016google}, board games~\citep{silver2017mastering}, and symbolic integration~\citep{lample2019deep}.

On the other hand, it has become increasingly evident that neural networks fall short in many aspects of human-level generalization, including those that symbolic approaches exhibit by design.
For example, it is difficult for neural language models to generalize syntactic rules such as verb tenses or embedded clauses in a systematic manner~\citep{keysers2020measuring,lake2018generalization,loula2018rearranging,hupkes2019compositionality}.
Similarly, in vision, neural approaches often learn overly specialized features that do not easily transfer to different datasets or held-out combinations of attributes~\citep{yosinski2014how,atzmon2016learning,santoro2018measuring}.
In reinforcement learning, where the use of neural networks has led to superhuman performance in gameplay~\citep{mnih2015humanlevel,silver2017mastering,berner2019dota}, it is found that agents are fragile under distributional shift~\citep{kansky2017schema,zhang2018study,gamrian2019transfer} and require substantially more training data than humans~\citep{tsividis2017human}.
These failures at systematically reusing knowledge suggest that neural networks do not learn a compositional knowledge representation (although some mitigation is possible~\citep{hill2019learning,hill2020environmental}).
In some cases, such as in vision, it may appear that object-level abstractions can emerge naturally as a byproduct of learning~\citep{zhou2014object}.
However, it has repeatedly been shown that such features are best understood as ``a texture detector highly correlated with an object''~\citep{olah2020zoom,sundararajan2017axiomatic,ancona2017unified,brendel2019approximating,geirhos2018imagenettrained}.
In general, evidence indicates that neural networks learn mostly about surface statistics (\eg between textures and classifications in images) in place of the underlying concepts~\citep{jo2017measuring,karpathy2015visualizing,lake2018generalization}.

A hybrid approach that combines the seemingly complementary strengths of neural networks and symbolic approaches may help address these issues, and several variations have been explored~\citep{bader2005dimensions}.  
A common variant uses a neural network as a perceptual interface (or pre-processor) tasked with learning symbols from raw data, which then serve as input to a symbolic reasoning system~\citep[\eg][]{mao2019neurosymbolic}.
Similarly, bottom-up neural networks have  been used to make inference more tractable in probabilistic generative models that contain the desired symbolic structure (\eg in the form of a symbolic graphics renderer~\citealp{kulkarni2015picture}).
Neural networks have also been combined with search-based methods to improve their efficiency~\citep{silver2016mastering}.
Countless other variations that vary in terms of the division of ur between the symbolic and neural components and the choice of a mechanism used to couple them are possible~\citep{mcgarry1999hybrid,davidson2020investigating}.

In this work, we will adopt a more unified approach that addresses these problems from within the framework of connectionism.
It is concerned with incorporating inductive biases in neural networks that enable them to efficiently learn about symbols and the processes for manipulating them (examples of such an approach are abound, even in early connectionist research, \eg \citet{smolensky1990tensor,pollack1990recursive,mcmillan1992rule,das1993unified}).
Compared to a hybrid approach, we believe that this is advantageous for a number of reasons.
Firstly, it reduces the required amount of task-specific engineering\footnote{
	This leaves the question of the innateness of aspects like causality or three-dimensional space open.
	Such priors might be helpful or eventually even necessary, however, an intelligent system must also be capable of independently discovering and using novel concepts and structures.
} and helps generalize to domains where expert knowledge is not available.
Secondly, by tightly integrating multiple different layers of abstraction, they can continuously co-adapt, which avoids the need for rigid interfaces between connectionist and explicitly symbolic components.
Finally, as is evident from the brain, it is sufficient to simply \emph{behave} as an emergent symbol manipulator, and therefore explicit symbolic structure is not a requirement. 
The main challenge regarding this approach to AI is then to identify corresponding inductive biases that enable symbolic behavior to emerge.

\subsection{The Binding Problem in Connectionist Methods}
\label{sec:bind:binding}

\begin{figure}[tb]
    \centering
    \includegraphics[width=\textwidth]{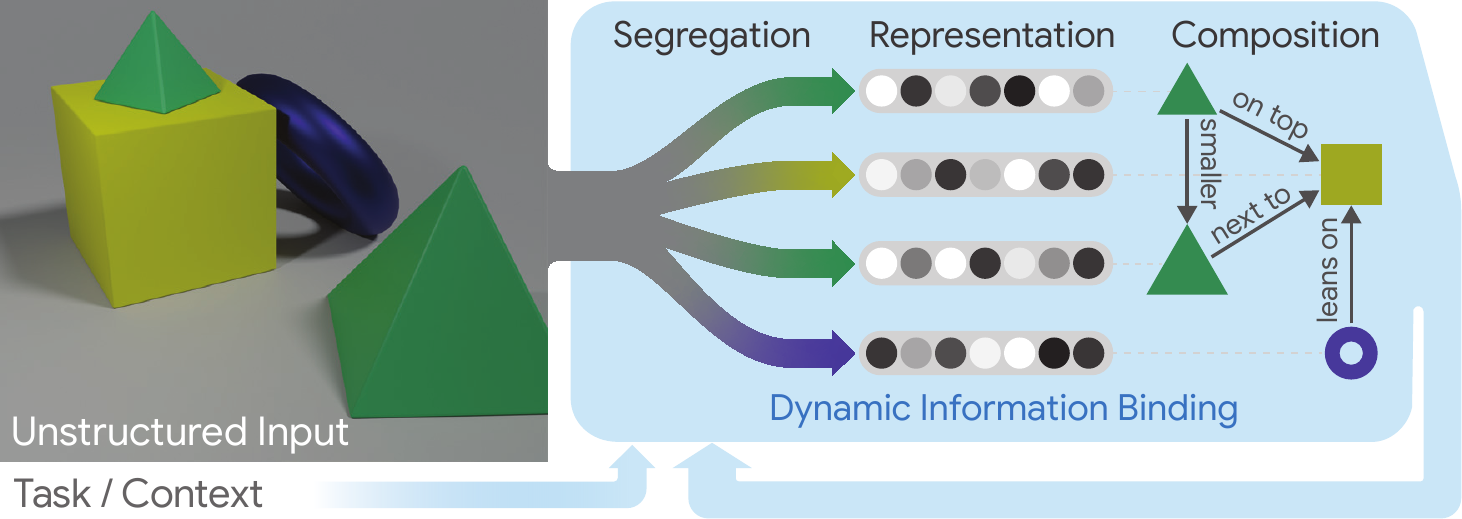}
    \caption{
    The binding problem in artificial neural networks can be understood from the perspectives of \emph{segregation}, \emph{representation}, and \emph{composition}.
    Each of these subproblems focuses on a different functional aspect of dynamically binding neurally processed information with the aim of facilitating more symbolic information processing.}
    \label{fig:bind:overview}
\end{figure}

We claim that there exists an underlying cause for the lack of emergent symbolic processing in neural networks, which we refer to as the binding problem.
The binding problem is about the inability to dynamically and flexibly combine (bind) information that is distributed throughout the network, which is required to effectively form, represent, and relate symbol-like entities.
In regular neural networks, information routing is largely determined by the architecture and weights, both of which are fixed at training time.
This limits their ability to dynamically route information based on a particular context and thereby accommodate different patterns of generalization.

The binding problem originates from neuroscience, where it is about the explanatory gap in our understanding of information processing in the brain.
It includes perceptual binding problems such as visual binding (color, shape, texture), auditory binding (a voice from a crowd), binding across time (motion), cross-modal binding (sound and vision into joint event), motor-behavior (an action), and sensorimotor binding (hand-eye coordination)~\citep{treisman1996binding,roskies1999binding,feldman2013neural}. 
Another class---sometimes referred to as cognitive binding problems---includes binding semantic knowledge to a percept, memory reconstruction, and variable binding in language and reasoning\footnote{The term binding problem has also been used in the context of consciousness, as the problem of how a single unitary experience arises from the distributed sensory impressions and processing in the brain~\citep{singer2001consciousness}}.

In the case of neural networks, the binding problem is not just a gap in understanding but rather characterizes a limitation of existing neural networks.
Hence, it poses a concrete implementation challenge to address the need for binding neurally processed information, which we believe is common to all of the above subproblems.
On the other hand, although we are convinced that this problem can be addressed by incorporating a general dynamic information binding mechanism, it is less clear how this can be implemented. 
Indeed, the search for an adequate mechanism for binding (in one form or another) is a long-standing problem, not just in neuroscience and cognitive psychology, but also in machine learning~\citep{smolensky1987analysis,smolensky1988proper,sun1992variable}.
Rather than focusing on a particular subproblem, here we propose to tackle the binding problem in its full generality, which touches upon all these related areas of research.
In this way, we can connect ideas from otherwise disjoint areas, and thus draw upon a large body of research towards developing a general binding mechanism.    
Inspired by \citet{treisman1999solutions}, we organize our analysis along a functional division into three aspects pertaining to the role of binding for symbolic information processing in neural networks: \emph{1) representation, 2) segregation, and 3) composition}, each of which takes a different perspective on the binding problem.

\paragraph{The Representation Problem} is concerned with binding together information at a representational level that belongs to separate symbol-like entities.
It revolves around so-called \emph{object representations}, which act as basic building blocks for neural processing to behave symbolically.
Like symbols, they are self-contained and separate from one another such that they can be related and assembled into structures without losing their integrity.
But unlike symbols, they retain the expressive distributed feature-based internal structure of connectionist representations, which are known to facilitate generalization~\citep{hinton1984distributed,bengio2013representation}.
Hence, object representations encode relevant information in a way that combines the richness of neural representations with the compositionality of symbols.
We chose the term ``object’’ representation because it is evocative of physical objects, which are processed as symbols in many important cognitive tasks.
However, we emphasize that object representations are also meant to encode non-visual entities such as spoken words, imagined or remembered entities, and even more abstract entities such as categories, concepts, behaviors, and goals\footnote{
We have considered several other terms for ``object’’ representations, including entity, gestalt, icon, and concept, which perhaps better reflect their abstract nature but are also less accessible at an intuitive level.
The fact that objects are more established in the relevant literature gave them the final edge.
}.

Interestingly, even the seemingly basic task of incorporating object representations in neural networks faces several problems, such as the ``superposition catastrophe''~\citep{vondermalsburg1986am} portrayed in \cref{fig:bind:superposition}.
It suggests that fully-connected neural networks suffer from an ``inherent tradeoff between distributed representations and systematic bindings among units of knowledge''~\citep{hummel1993distributing}.
A general treatment of object representation in neural networks involves addressing the superposition catastrophe, along with several other challenges, which we discuss in \cref{sec:rep}.

\begin{figure}[t]
    \centering
    \includegraphics[width=\textwidth]{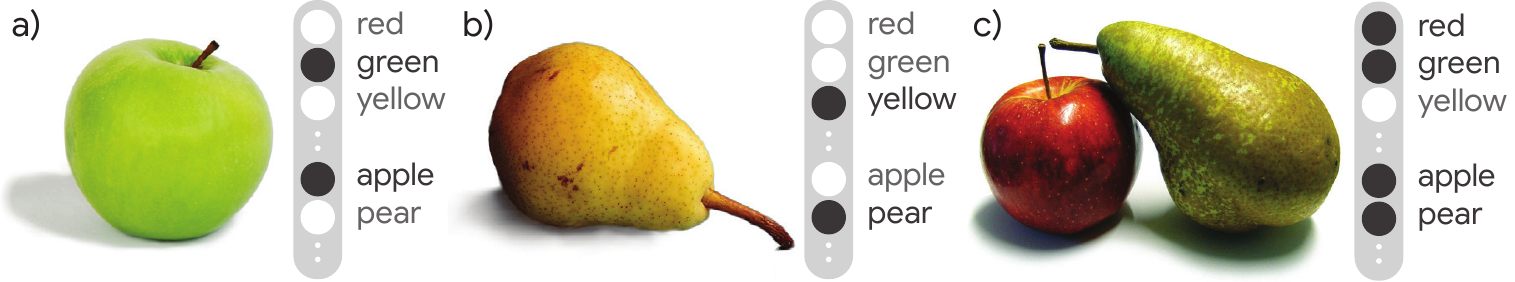}
    \caption{Illustration of the \emph{superposition catastrophe}: 
    A distributed representation in terms of disentangled features like color and shape \textbf{(a, b)} leads to ambiguity when confronted with multiple objects \textbf{(c)}:
    The representation in \textbf{(c)} could equally stand for a red apple and a green pear, or a green apple and a red pear.
    It leads to an indiscriminate bag of features because there is no association of features to objects.
    A simple form of this problem in neural networks was first pointed out in \citet{rosenblatt1961principles}, and has been debated in the context of neuroscience since~\citep{milner1974model,vondermalsburg1981correlation}.}
    \label{fig:bind:superposition}
\end{figure}

\paragraph{The Segregation Problem} is about the process of structuring raw sensory information into meaningful entities.
It is concerned with the information binding required for dynamically \emph{creating} object representations, as well as the characteristics of objects as modular building blocks for guiding this process.
This notion of an object is context and task-dependent, and difficult to formalize even for concrete objects like a tree, a hole, or a river, which are self-evident to humans.  
Hence, the segregation problem relates to the problem of instance segmentation in that it also produces a division of the input into meaningful parts, but it is complicated by the fact that it is concerned with objects in their most general form.
The incredible variability among objects makes it intractable to resolve the segregation problem purely through supervision.
Consequently, the segregation problem (\cref{sec:seg}) is about enabling neural networks to acquire an appropriate, context-dependent, notion of objects in a mostly unsupervised fashion.

\paragraph{The Composition Problem} is about using object representations to dynamically construct compositional models for inference, prediction, and behavior.
These structured models leverage the modularity of objects to support different patterns of generalization, and are the means by which more systematic `human-like’ generalization can be accomplished. 
However, this relies on the ability to learn abstract relations that can be arbitrarily and recursively applied to object representations, and requires a form of binding, not unlike the way variables can be bound to placeholder symbols in a mathematical expression.
Moreover, the desired structure is often not known in advance and has to be inferred or adapted to a given context or task.
To address the composition problem (\cref{sec:comp}), a neural network thus requires a mechanism that provides the flexibility to quickly restructure its information flow and ultimately enable it to generalize systematically.

}

{
\section{Representation}
\label{sec:rep}

\begin{figure}
    \centering
    \includegraphics[width=0.95\textwidth]{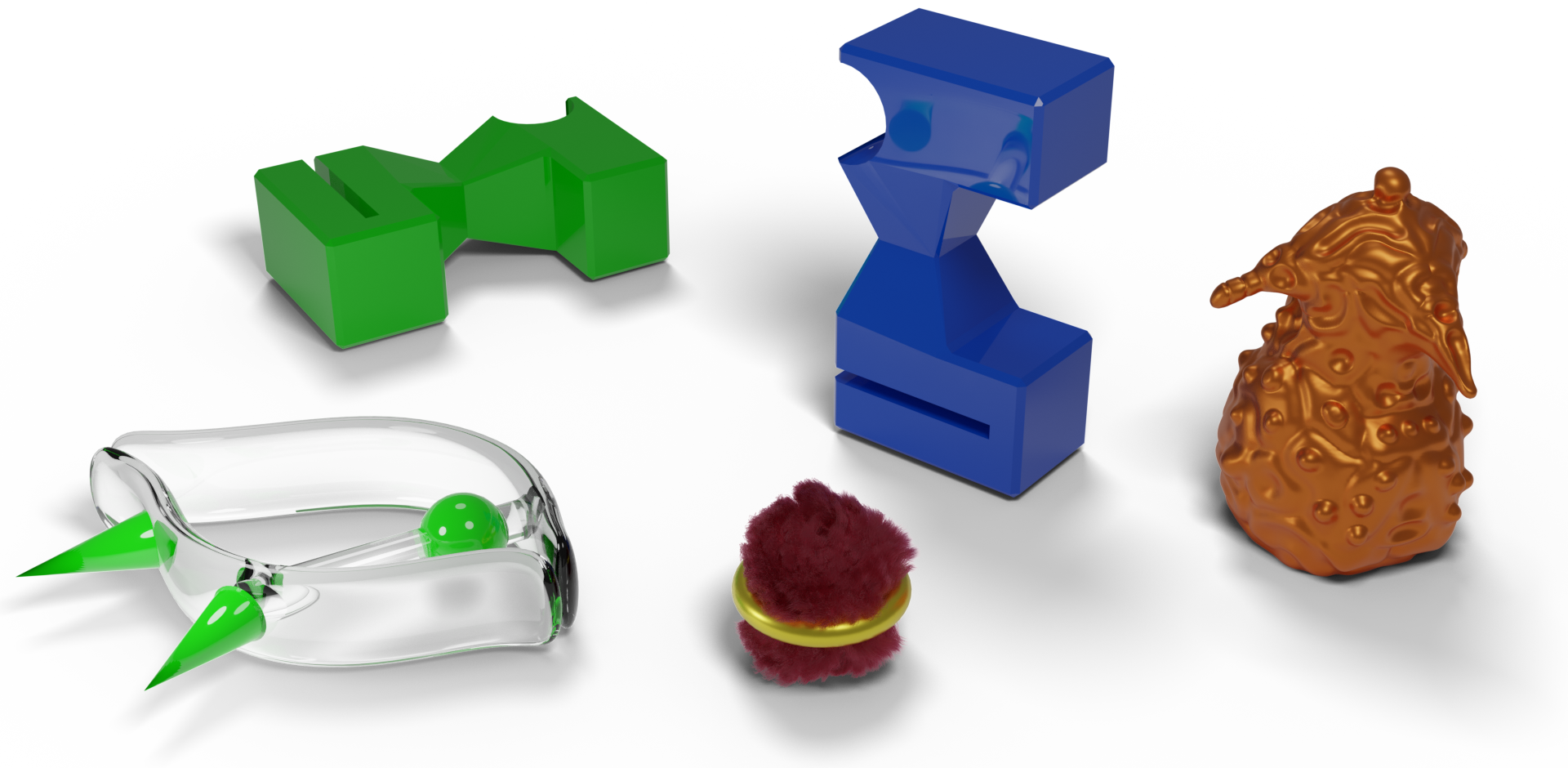}
  	\caption{A visual scene composed of various unfamiliar objects.}
  	\label{fig:rep:multi_object}
\end{figure}

In this section, we look at the binding problem from the perspective of representation.
We have argued that, to take advantage of symbolic processing, neural networks require some form of object representations that combine the richness of neural representations with the compositionality of symbols.
These object representations are intended as modular `building blocks’ from which to efficiently compose structured models of the world.
This has direct consequences for the representational format and its underlying dynamics.

Consider for example \cref{fig:rep:multi_object}, where you are able to distinguish between five different objects.
You can readily describe each object in terms of its shape, color, material, and other properties, despite most likely never having encountered them before.
Notice also how these properties relate to individual objects as opposed to the entire scene, which is also evident from the fact that you can tell that the color green occurs multiple times for different objects.
Finally, notice how you are readily able to perform comparisons, for example, to tell that the shape of the blue object is the same as that of the green one in the back, but that they differ in color.

In the following, we take a closer look at the \emph{format} of object representations (\cref{sec:rep:format}).
We work towards a format that separates information about objects and is general enough to accommodate unfamiliar objects in a meaningful way so that they can readily be compared.
Additionally, we will also consider the representational \emph{dynamics} that are required to support stable and coherent object representations over time (\cref{sec:rep:dynamics}).  
Towards the end, we survey relevant approaches from the literature that may help incorporate these aspects of object representations into neural networks (\cref{sec:rep:methods}).

\subsection{Representational Format}
\label{sec:rep:format}

\begin{figure}
    \centering
    \includegraphics[width=\textwidth]{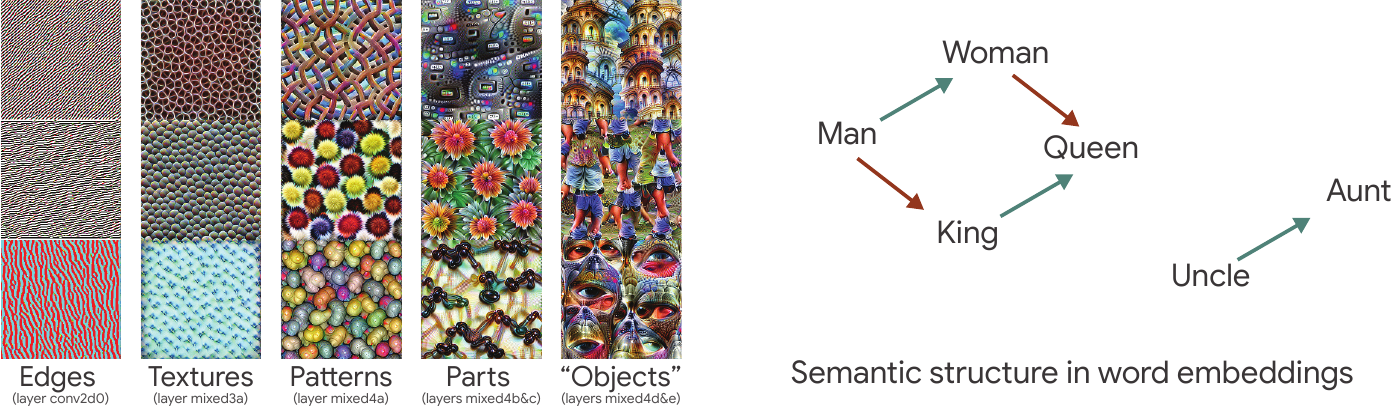}
    \caption{
        \textbf{Left:} Interpretable features learned on ImageNet as observed in~\citet{olah2017feature}. 
        \textbf{Right:} Learned word embeddings have been demonstrated to capture some of the semantic structure of text~\citep{mikolov2013efficient}, although to a lesser extent than was initially reported~\citep{nissim2019fair}.
      }
  \label{fig:rep:embeddings}
\end{figure}

We seek a representational format that distinguishes objects, while retaining the advantages of learned distributed representations.
These representations have proven highly successful \citep[\eg][]{ciresan2011flexible,hinton2012deep,krizhevsky2012imagenet} and are known to partially capture the semantic structure of a task (\cref{fig:rep:embeddings}), such as interpretable image features~\citep{zeiler2014visualizing,olah2020zoom}, or the semantic structure of text~(\citealp{mikolov2013efficient}; but compare \citealp{nissim2019fair}).
In this way learned object representations can also benefit from known inductive biases that focus on feature hierarchies, invariances, and spatio-temporal coherence~\citep{becker1992selforganizing}, sparsity~\citep{olshausen1996emergence}, or non-Euclidean feature spaces~\citep{nickel2017poincare}.

\subsubsection{Separation}
\label{sec:rep:separation}

To support the construction of structured models, object representations need to act as modular building blocks.
This requires information about individual objects to remain separated at a representational level, such that their features do not interfere with one another, even when composed.
Additionally, the features that belong to an object must be able to act as a unit, which implies strong dependencies between its features.
For example, when an object representation appears or ceases to exist, all of its features are equally affected.

The separation of information has to be flexible enough to ensure that objects can be formed from novel (unseen) feature combinations.
Hence, it is important that it is not purely determined by the representational content of the objects, but rather acts as an independent degree of freedom.
Regarding capacity, it may suffice to represent only a few objects simultaneously, despite the fact that a typical scene potentially contains a large number of objects.
Indeed, the capacity of the human working memory is generally believed to only be around 3--9 objects~\citep{fukuda2010discrete,miller1956magical}.

\subsubsection{Common Format}
\label{sec:rep:common_format}

To be able to efficiently relate and compare a wide variety of object representations, they must be described in a common format.
Recall how in \cref{fig:rep:multi_object} you were able to freely compare a number of unfamiliar objects in terms of their properties, such as their size, shape, and location.
On the one hand, this is possible because you have acquired a number of general relationships, such as ``bigger than'', ``left of'', \etc, which we will discuss in detail in \cref{sec:comp}.
What is more important here is that such relations can only be applied if object representations provide a shared interface.
More generally, a common format helps to ensure that \emph{any} learned relation, transformation, or skill (like grasping) transfers between similar objects independent of context. 
Similarly, a common set of features helps carry over experiences between objects during learning.

\subsubsection{Disentanglement}
\label{sec:rep:disentanglement}

Individual object representations need to be able to describe a large variety of (possibly unseen) objects in terms of attributes that are useful for down-stream problem-solving.
This requires focusing on factors of variation in the data, that are sufficiently expressive, but also compact and reusable (\ie they can be varied independently).
Indeed, humans arguably manage to accomplish this by focusing on a relatively small, but consistent set of attributes such as color, shape, \etc~\citep{devereux2014centre}.

A \emph{disentangled} representation aims to make these attributes explicit by establishing a local correspondence between (independent) factors of variation and features~\citep{barlow1989finding,schmidhuber1992learningb,higgins2017betavae,higgins2018definition,ridgeway2018learning}. 
In this case, information about a specific factor can be readily accessed and is robust to unrelated changes in the input, which improves sample efficiency and down-stream generalization~\citep{higgins2017darla,vansteenkiste2019are}.
In the context of object representations, disentanglement implies a factorized feature space that captures salient properties of objects.
Together with a common format, it facilitates generalization to unseen feature combinations and enables useful comparisons between objects and other meaningful relations to be formed.

\subsection{Representational Dynamics}
\label{sec:rep:dynamics}

When interacting with the real world, the stream of sensory information continuously evolves over time.
It is therefore important to consider not only instantaneous representations, but also their \emph{dynamics} over time.

\subsubsection{Temporal Dynamics}
\label{sec:rep:temporal}
An object representation requires ongoing updates across time for a number of reasons:
Firstly, with objects constantly moving and transforming in the real world, their corresponding representations need adjustments to remain accurate.
Secondly, certain temporal attributes such as movement or behavior can only be estimated when considering the history of information.
Finally, with the limited amount of information that can be observed about an object at any given time, accumulating information over multiple partial views can help produce more informative object representations.

An important aspect among all these cases is the need for an object representation to consider not only the input but also its own history (recurrence).
This requires a stable identity to help ensure that information across time-steps is associated with the correct object representation. 
Note that the identity of an object cannot be tied exclusively to its visible properties, as illustrated by the extreme example of a fairytale prince that is transformed into a frog~\citep{marcus2003algebraic,bambini2012conversation}.

\subsubsection{Reliability}
\label{sec:rep:reliability}

Structured mental models depend on object representations to provide a stable foundation for reasoning and other types of information processing~\citep{johnson-laird2010mental}.
The reliability of this foundation is especially important for more abstract computations to which object representations provide the only connection to the world.
However, perfect reliability is unattainable since sensory information about the world is noisy and incomplete, and the capacity of any model is inherently limited.

Explicitly quantifying uncertainty can help mitigate this issue and prevent noise and errors from accumulating undetectably.
In addition, certain small amounts of noise in an object representation may be continually corrected by leveraging dependencies among its features (\ie through the features of an object acting as a unit).
An important source of uncertainty accumulation is due to objects that are temporarily not perceived (\eg as a result of occlusion).
In this case, a `self-correcting’ representation may help maintain a stable object representation, even in the absence of sensory input (object permanence).

Uncertainty about object representations may also arise due to ambiguous inputs that allow for several distinct but coherent interpretations (for example see \vref{fig:rep:attractor}).
The ability to (at least implicitly) encode multi-modal uncertainty is crucial to effectively treat such cases.
Top-down feedback may then help disambiguate different interpretations (see also \cref{sec:seg:top_down,sec:comp:inferring_structure}).

\subsection{Methods}
\label{sec:rep:methods}

In order to fulfill the desiderata outlined above, we require a number of specialized inductive biases.
Indeed, it should now also be clear that a simple MLP falls short at adequately representing multiple objects simultaneously:
If it attempts to avoid the superposition catastrophe by learning features that are specific to each object, then they lack a common format and become difficult to compare\footnote{
Others have suggested ways in which MLPs could \emph{in principle} circumvent this problem \citep{oreilly2002generalizable, pollack1990recursive}.
However, neither of these offer a solution that can convincingly fulfill all of the above desiderata simultaneously.
In fact, even for plain RNNs it was found that when they are trained to remember multiple objects internally, they resort to a localist representation~\citep{bowers2014neural}. 
}.
Therefore, in the following we will review several approaches for representing multiple objects in neural networks.
We will focus on common format, temporal dynamics, reliability, and in particular on separation, which thus far has received little attention in the main-stream neural networks literature.

\subsubsection{Slots}
\label{sec:rep:slots}

\begin{figure}[ht]
    \centering
    \includegraphics[width=\textwidth]{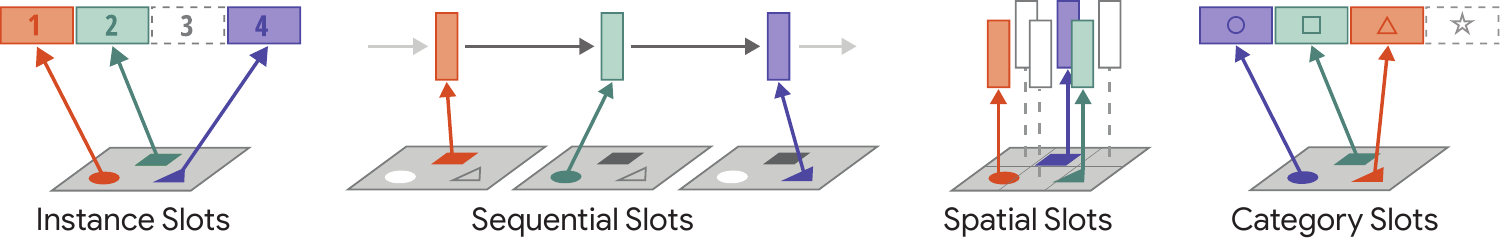}
    \caption{
        Illustration of the four different types of slot-based representations.}
    \label{fig:rep:slots}
\end{figure}

The simplest approach to separation is to provide a separate representational slot for each object. 
This provides a (typically) fixed capacity working memory with independent object representations that can all be accessed simultaneously.
Weight sharing can then be used to ensure a common format among the individual slots.

\paragraph{Instance Slots}
In the most general form, which we call \emph{instance slots}, all slots share a common format and their information can be kept separate, independent of their representational content.
Instance slots are very flexible and general in that they have no preference for content or ordering.
However, this generality introduces a \emph{routing problem} when a common format is enforced via weight sharing: with all slots being identical, bottom-up information processing needs to break this symmetry to avoid assigning the same content to each one.
Hence, the allocation of information to each slot must be determined by taking the other slots into account, which complicates the process of segregation (see also \cref{sec:seg:dynamics}).
Instance slots have been used in several approaches to learning object representations, including Masked Restricted Boltzman Machines (M-RBMs;~\citealp{leroux2011learning}), Neural Expectation-Maximization~(N-EM;~\citealp{greff2017neural}), and IODINE~\citep{greff2019multiobject}.
They can also be found in the memory of memory-augmented neural networks~\citep{joulin2015inferring,graves2016hybrid}, in self-attention models~\citep{vaswani2017attention,dehghani2019universal,locatello2020objectcentric}, in Recurrent Independent Mechanisms (RIMs;~\citep{goyal2019recurrent}), albeit without having a common format, and in certain graph neural networks~\citep{battaglia2018relational}, where they are treated as internal representations that can be accessed simultaneously.

\paragraph{Sequential Slots}
Sequential slots break slot symmetries by imposing an order on the representational slots, typically across time.
They are commonly found in RNNs and, when paired with an attention mechanism that attends to a different object at each step, can serve as object representations.
With weights typically being shared across (time)steps, sequential slots naturally share a common format and unlike other slot-based representations can dynamically adjust their representational capacity.
Sequential slots in RNNs have been used as object representations, for example in Attend Infer Repeat (AIR;~\citealp{eslami2016attend}) and to a lesser degree in DRAW~\citep{gregor2015draw}.
However, due to recurrence, these slots may not always be fully independent, which impedes their function as modular building blocks.
Recent approaches, such as Multi-Object Networks (MONet;~\citealp{burgess2019monet}) and GENESIS~\citep{engelcke2019genesis}, alleviate this by using recurrence only for information routing, but not for the object representations themselves.
In general, a potential limitation of sequential slots is that they are not simultaneously accessible at any given (time)step for down-stream processing. 
This can be addressed via a set function over sequential slots, such as the attention mechanism in certain neural machine translation methods~\citep{bahdanau2014neural} or in pointer networks~\citep{vinyals2015pointer}.

\paragraph{Spatial Slots}
In \emph{spatial slots}, each slot is associated with a particular spatial coordinate (\eg in an image), which helps to break slot symmetries and simplifies information routing.
They can still accommodate a common format through weight-sharing, but lack generality because their content is tied to a specific spatial location.
Because location and separation are entangled, changes to the location of an object potentially correspond to a change of slot, which complicates maintaining object identity across time.
Spatial slots are commonly found in CNNs, where multiple convolutional layers share filter weights across the spatial dimensions to yield a spatial map of representational slots.
Although they are not usually advertised as object representations in this way, several recent approaches, such as Relation Networks~\citep{santoro2017simple}, the Multi-Entity VAE~\citep{nash2017multientity}, or the works by \citet{zambaldi2019deep,stanic2020hierarchical} explicitly treat each spatial position in the filter-map of a CNN as a candidate object representation.
Even more recent approaches, such as SPAIR~\citep{crawford2019spatially}, SPACE~\citep{lin2020space}, and SCALOR~\citep{jiang2020scalor}, expand on this by incorporating explicit features for the presence of an object and its bounding box into each spatial slot.
Nonetheless, a current limitation of these approaches is that their spatial slots are typically tailored towards objects that are reasonably well separated, and whose size is compatible with the corresponding receptive field (or the bounding box) in the image.

\paragraph{Category Slots}
A related approach is to allocate slots according to some categorization of objects based on properties other than location.
This too can serve to break slot symmetries for the purpose of information routing, and is further expected to mitigate the dependence of spatial slots on spatially separated inputs.
In this case, however, because now category and separation are entangled, it is then no longer possible to represent multiple objects of the same category\footnote{\label{foot:rep:overlap}
There is some evidence that humans struggle with feature overlap too and show reduced working memory capacity in these cases~\citep{mozer1989types}.}.
The main example of category slots are capsules~\citep{hinton2011transforming, hinton2018matrix}, although other approaches such as Recurrent Entity Networks~\citep{henaff2016tracking} can also be viewed from this perspective.

\subsubsection{Augmentation}
\label{sec:rep:augmentation}

\begin{SCfigure}
    \centering
    \includegraphics[width=0.5\textwidth]{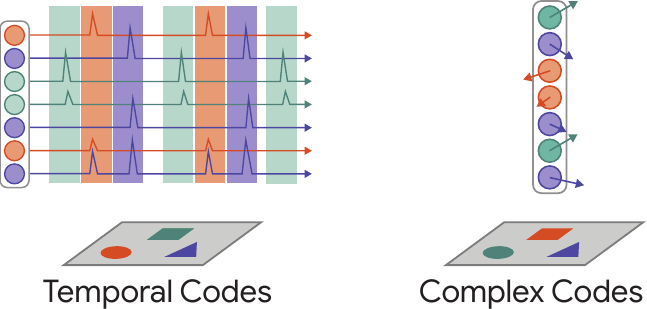}
    \caption{
        Illustration of the two main augmentation based approaches to object representations. 
\textbf{Left:} Neural activity over time for a temporal code, where synchronization is emphasized using color. 
\textbf{Right:} Complex valued activations are represented by arrows and colored according to their direction.
}
    \label{fig:rep:augment}
\end{SCfigure}

Augmentation based approaches, unlike slot based ones, keep a single set of features shared among all object representations and instead augment each feature with additional grouping information.
This grouping information is usually continuous, which may help to encode uncertainty about the separation.
Object representations based on augmentation will trivially be in a common format, although extracting information about individual objects now requires first processing the grouping information.
An important limitation of augmentation is that it requires substantial deviations from standard connectionist systems and is thus more difficult to integrate with state of the art systems.
Due to features being shared, augmentation may also suffer from capacity and ambiguity problems when a feature is active in multiple object representations at the same time (\eg two red objects), similar to when representing multiple objects of the same category using category slots~\cref{foot:rep:overlap}.

\paragraph{Temporal Codes}
An early approach to object representation using augmentation in neural networks made use of the temporal structure of \emph{spiking neurons} for separation (temporal codes).
Here, the activation of a feature encoded by the firing rate is augmented with grouping information encoded by the temporal correlation between firing patterns~\citep{singer2009distributed}.
In other words, the features that form an object are represented by neurons that fire in synchrony~(\citealp{milner1974model,vondermalsburg1981correlation,singer1999neuronal}; see also \cref{sec:rel:binding_problem}).
Rather than using unrestricted spiking networks, most work on object representation using temporal codes focuses on \emph{oscillatory networks}, where the firing pattern takes the form of a regular frequency rhythm (for an overview see \cite{wang2005time}).
Because temporal codes rely on spiking neurons, they are non-differentiable and also require simulating the dynamics of each neuron even for static inputs.
This makes them incompatible with gradient-based training, and necessitates a completely different training framework~\citep[\eg][]{doumas2008theory,doumas2019relation} typically based on Hebb's rule~\citep{kempter1999hebbian}, or Spike-Timing-Dependent Plasticity (STDP;~\citealp{caporale2008spike}).

\paragraph{Complex-Valued Codes}
An alternative approach to augmentation uses \emph{complex-valued} neurons (features) in place of oscillatory neurons.
Hence, instead of explicitly simulating the temporal behavior of an oscillator, its activation and grouping information can now be described as the absolute value and angle of a complex-valued neuron.
Similar to before, the grouping is implicit and smooth with neurons that ``fire at similar angles’’ being grouped together.
Complex-valued neurons are differentiable and more compatible with existing gradient-based learning techniques.
On the other hand, they require specialized activation functions that consider both real and imaginary parts\footnote{In some sense, complex codes can be seen as an instance of a more general – yet unexplored – class of vector-valued activations that use the additional degrees of freedom for grouping.}, which tend to be difficult to integrate with existing methods.
Successful integrations include complex-valued Boltzmann Machines~\citep{reichert2013neuronal, zemel1995lending} and complex-valued RNNs that could be trained either with backpropagation~\citep{mozer1992learning} or via Hebbian learning~\citep{rao2008unsupervised}.

\subsubsection{Tensor Product Representations}
\label{sec:rep:tpr}

\begin{SCfigure}
    \centering
    \includegraphics[width=0.4\textwidth]{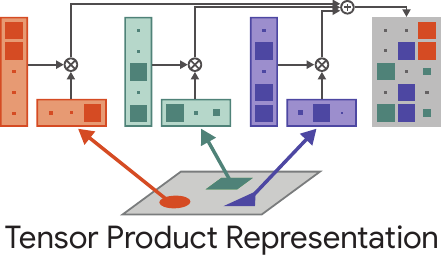}
    \caption{
        Illustration of a Tensor Product Representation (matrix on the right) that is formed through combining a role vector (horizontal) and a filler vector (vertical) for each object.}
    \label{fig:rep:tpr}
\end{SCfigure}

A Tensor Product Representation (TPR) consists of a real-valued matrix (tensor) that is the result of combining distributed representations of \emph{fillers} with distributed representations of \emph{roles}.
TPRs can be used for representing multiple objects by associating fillers with object representations and using roles to encode grouping information.
A TPR is formed by combining each filler with a corresponding role via an outer product (``binding operation’’), which are then composed to accommodate multiple object representations (``conjunction operation’’).
When the role representations are linearly independent, then the object representations can be retrieved from the TPR via matrix multiplication (``unbinding operation’’).
Notice that, when the role-vectors are one-hot encodings, the TPR reduces to instance slots. 
However, the additional freedom afforded by a general \emph{distributed} role vector can be used to encode structural information or uncertainty about the separation of objects.
TPRs always assume that the object representations are described in a common format.
But note that, similar to augmentation, extracting information about individual objects first requires processing the grouping information (in this case via the unbinding operation).
TPRs were first introduced in \cite{smolensky1990tensor} and several modifications have since been proposed that consider different binding, unbinding, and conjunction operations~(\citealp{plate1995holographic,kanerva1996binary, gayler1998multiplicative}; see \citealp{kelly2013encoding} for an overview).
In the recent literature, TPR-like mechanisms have been incorporated into neural networks using fast-weights~\citep{schlag2018learning} or self-attention~\citep{schlag2019enhancing} to perform reasoning in language.

\subsubsection{Attractor Dynamics}
\label{subsec:rep:attractor}

\begin{SCfigure}
    \centering
    \includegraphics[width=0.7\textwidth]{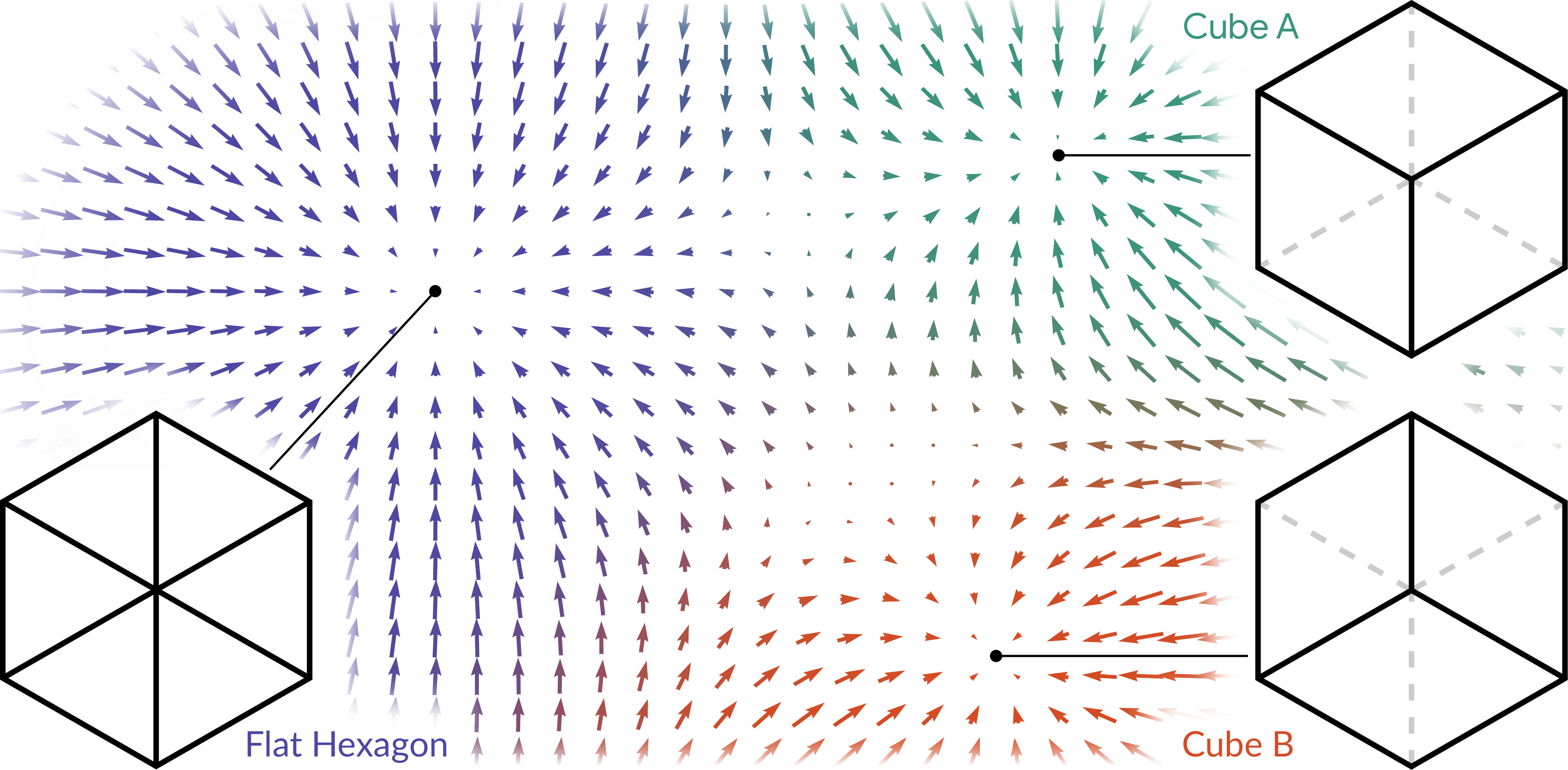}
    \caption{
        Correspondence of attractor states to visual interpretations for a tri-stable variant of the Necker cube.
        The vector field illustrates the (input-dependent) inference dynamics in feature space, with one attractor for each stable interpretation.
    }
    \label{fig:rep:attractor}
\end{SCfigure}

Up until this point, we have focused on methods that address the representational format of object representations.
Now we consider \emph{attractor dynamics} as an approach for addressing their representational dynamics (\cref{sec:rep:dynamics}).
Robust object representations are well described by a stable attractor state in a larger dynamical system that models the representational dynamics based on a given input.
In this case, inferring a coherent object representation corresponds to running the dynamical system forward until it converges to an attractor state.
A stable attractor is naturally self-correcting, and multiple competing interpretations (from ambiguous inputs) can easily be described by separate attractor states.
Top-down feedback can then be used to switch interpretations by pushing the state of the system enough to cause it to cross over to a different basin of attraction.
By adapting the system dynamics to changing inputs, they allow for moving attractors (changes of the object) or bifurcations (creation or vanishing of interpretations).

Attractor Networks incorporate attractor dynamics in neural networks and have a long history in connectionist research.
Early work includes Hopfield networks~\citep{hopfield1982neural}, Boltzmann machines~\citep{ackley1985learning}, and associative memory~\citep{kohonen1989selforganization}.
Attractor states were also found to occur naturally in RNNs, especially when using symmetric recurrent weights~\citep{almeida1987learning, pineda1987generalization}.
In recent years, however, they have received little attention (but see~\cite{mozer2018statedenoised,iuzzolino2019convolutional}), which might be in part because they can be difficult to train.
In particular, the fact that each weight participates in the specification of many attractors can lead to spurious (unintended) attractors and ill-conditioned attraction basins~\citep{neto1999multivalley}.
Localist attractor networks~\citep{zemel2001localist} and flexible kernel memory~\citep{nowicki2010flexible} are two approaches that address this issue by introducing a separate representation for each attractor.
However, note that spurious attractors that correspond to novel feature combinations may also be advantageous for generalization.

\subsection{Learning and Evaluation}
\label{sec:rep:summary}

Object representations are the product of segregation and the foundation upon which compositional reasoning is built.
To effectively connect high-level abstract reasoning with low-level sensory data they must be learned jointly, together with composition and segregation.
Learning object representations requires incorporating architectural inductive biases to ensure a common format and to provide enough flexibility for dynamically separating information.
Regarding separation, slot-based approaches offer a simple and minimal approach, while augmentation and TPRs are more difficult to incorporate, yet support more sophisticated use cases.
The problem of learning representations that are disentangled can be approached by optimizing for some notion of (statistical) independence between features~\citep[\eg][]{schmidhuber1992learningb,chen2016infogan,higgins2017betavae}, sparse feature updates across time~\citep{whitney2016disentangled}, or independent controllability of features~\citep{thomas2017independently}.
In terms of temporal dynamics and robustness, the situation is less clear, although the use of attractor networks may serve as a good starting point. 

Evaluation plays a critical role in guiding research to make measurable progress towards good object representations. 
A useful approach is to measure how well the system copes with particular generalization regimes such as to held-out-combinations of features for disentanglement~\citep{esmaeili2018structured} and separation~\citep{santoro2018measuring}, prediction roll-outs for temporal dynamics~\citep{vansteenkiste2018relational}, and robustness to injected noise for reliability~\citep{mozer2018statedenoised}.
However, in case of poor performance it may be difficult to diagnose the source of the problem in terms of properties of the representational format and dynamics.
When ground truth information is available, an alternative is to directly measure selected properties of the object representations, such as local correspondence between ground-truth factors of variation and features for disentanglement~\citep{eastwood2018framework}.
Finally, qualitative measures such as latent traversals or projections of the embedding space~\citep{maaten2008visualizing} can provide an intuition about the learned representations but due to their subjectivity, quantitative measures should be preferred.

}
{
\section{Segregation}
\label{sec:seg}

\begin{figure}
  \centering
  \includegraphics[width=\textwidth]{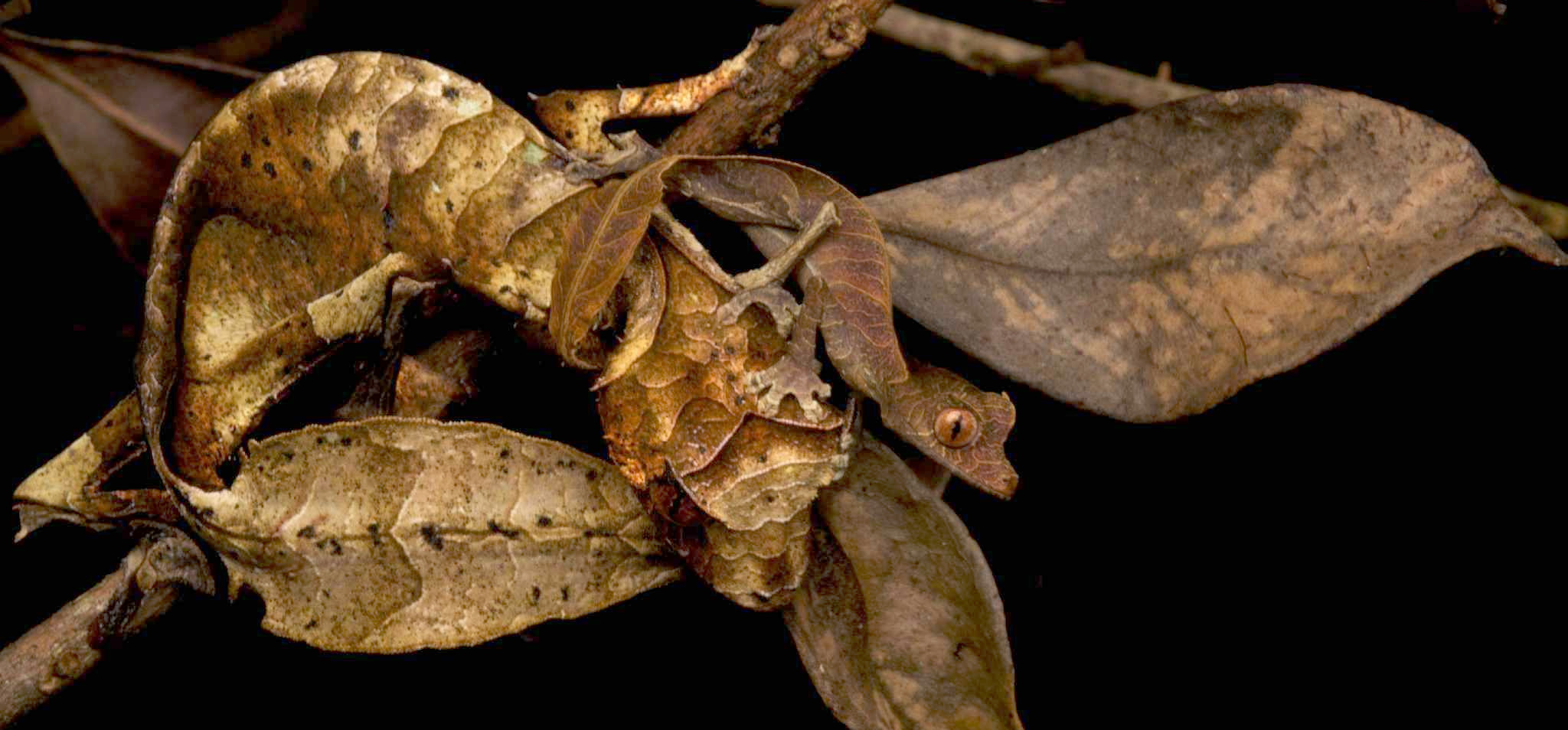}
  \caption{Photo of two leaf-tailed geckos --- “young and old” \raisebox{1.5pt}{\scriptsize\textcopyright} 2015 by Paul Bertner.}
  \label{fig:seg:gecko}
\end{figure}

In this section, we look at the binding problem from the perspective of segregation: the process of forming object representations.
Unlike in \cref{sec:rep}, where we focused on the need for binding at a representational level to maintain a separation of information for \emph{given} entities, here we focus on the process of \emph{creating} object representations through binding previously unstructured (raw) sensory information.
Humans effortlessly perceive the world in terms of objects, yet this process of perceptual organization is surprisingly intricate~\citep{wagemans2015oxford}. 
Even for everyday objects like a mirror, a river, or a house, it is difficult to formulate precise boundaries or a definition that generalizes across multiple different contexts.
Nonetheless, we argue that an important aspect common to all objects is that they may act as stable and self-contained abstractions of the raw input.
This then has important implications for the process of segregation.

Consider for example \cref{fig:seg:gecko}, which demonstrates several challenges for segregation that must be overcome.
To recognize the two geckos sitting on a branch you have to segment out two unfamiliar objects (zero-shot) even though they belong to the same class (instance segmentation) and their use of camouflage (texture similarity).
Both the large gecko and the branch are visually disconnected due to occlusion, and yet you perceive them as independent wholes (amodal completion).
Beyond separating these objects, you have also formed separate representations for them that enable you to efficiently relate, describe, and reason about them.

In the following,, we take a closer look at this process of segregation\footnote{We refer to this process as \emph{segregation} rather than \emph{binding}, to emphasize the fact that it typically requires a \emph{separation} of the inputs and features into meaningful parts.}.
We first work towards a general \emph{notion} of an object built around modularity and hierarchy (\cref{sec:seg:objects}).
Next, we focus on the process of \emph{forming} object representations based on this notion (\cref{sec:seg:dynamics}).
Unlike segmentation, which is typically only concerned with a static split at the \emph{input}-level, segregation is inherently task-dependent and aims to produce stable object representations that are grounded in the input and which maintain their identity over time.
Towards the end, we survey relevant approaches from the literature that may help neural networks perform segregation (\cref{sec:seg:methods}).

\subsection{Objects}
\label{sec:seg:objects}

The question of what constitutes a meaningful object (\ie for building structured models of the world) is central to segregation.
However, despite long-standing debates in many fields including philosophy, linguistics, and psychology, there exists no general agreed-upon definition of objects~\citep{green2018theory,cantwell-smith1998origin}.
Here, we take a pragmatic stance that focuses on the \emph{functional role} of objects as compositional building blocks.
Hence, we are not interested in debating the ``true’’ (\ie metaphysical) nature of objects, but rather consider object representations as components of a useful representational ``map'' that refers to (but is not identical to) parts of the ``territory’’ (world)\footnote{\textit{``A map is not the territory it represents, but, if correct, it has a similar structure to the territory, which accounts for its usefulness.''}~\citep{korzybski1958science}.}.

\subsubsection{Modularity}
\label{sec:seg:modularity}

\begin{SCfigure}
  \centering
  \includegraphics[width=0.7\textwidth]{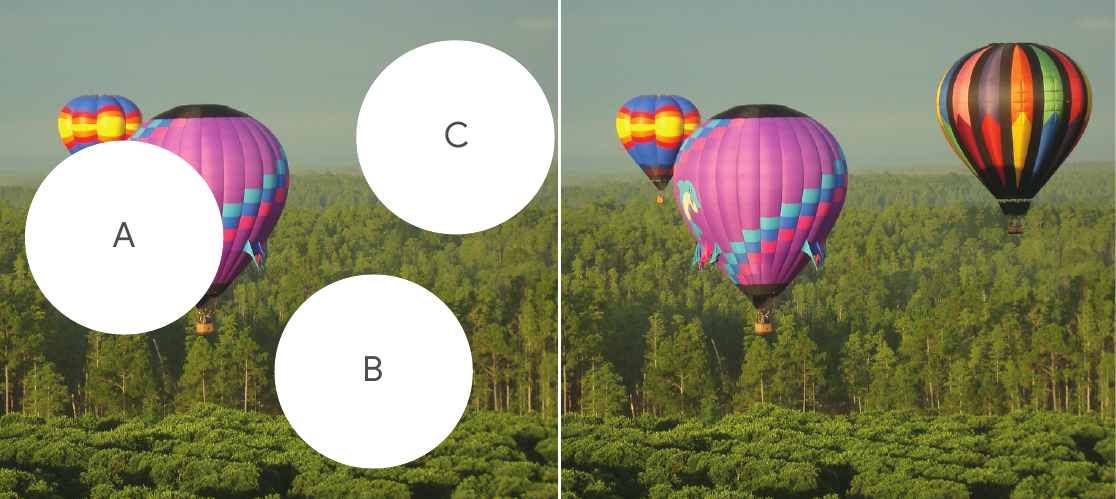}
  \caption{For partial objects \textbf{(A)} or only background \textbf{(B)}, the occluded regions can be inpainted reasonably well, while in the case of full object occlusion \textbf{(C)} that is usually impossible.}
  \label{fig:seg:balloons}
\end{SCfigure}
From a functional perspective, the defining quality of an object is that it is modular, \ie it is self-contained and reusable independent of context.
While this suggests choosing objects with minimal information content (to improve reusability), it is equally important that objects can be represented efficiently based on their internal predictive structure. 
We argue that this trade-off induces a Pareto front of valid decompositions into objects that have both strong internal structure, yet remain largely independent of their surroundings.
By organizing information in this way, objects are expected to capture information that is due to independent causes, which matches our intuitive notion of objects in the real world~\citep{green2018theory,chater1996reconciling}.

Consider the example of three balloons in front of a forest as depicted in \cref{fig:seg:balloons}.
When a balloon is partially occluded (as in A), you are still able to make a reasonable guess about the occluded part purely based on its internal predictive structure.
On the other hand, when an entire balloon is occluded (as in B) it is impossible to infer its presence from the (unoccluded) context, and the most reasonable reconstruction is to fill in based on the background (as in C).
Notice that each balloon is modular in the sense that it is possible to reuse them in many different contexts (\eg when placed in a different scene). 
In contrast, this would not be possible if an object were to be formed from the background \emph{and} the balloon.
Hence, by carving up perception at the ``borders of predictability'', objects allow for an approximate divide and conquer (\ie a compositional) approach to modeling the world.

\subsubsection{Hierarchical}
\label{sec:seg:hierachical}
Objects are often hierarchical in the sense that they are composed of parts that can themselves be viewed as objects. 
Consider, for example, a house consisting of a roof and walls, which themselves may consist of several windows and a door, etc.
Depending on the desired level of detail, a scene can therefore be decomposed in terms of coarser or finer scale objects, corresponding to different solutions on the Pareto front.
In most cases, these decompositions relate to each other in the sense that they correspond to different levels in the \emph{same} part-whole hierarchy.
However, in rare cases, two decompositions may also consider incompatible parts, as, for example, in a page of text that can be decomposed either into lines or sentences\footnote{
A unique hierarchy is favored by modularity because in the case of incompatible decompositions (\ie not corresponding to the same part-whole hierarchy) their objects \emph{cross} “borders of predictability”, which implies a weaker internal structure.
}.
Notice that there is a difference between this part-whole hierarchy and the feature hierarchy typically found in neural networks.
Here, parts are themselves objects, which are the result of dynamically separating information into object representations (segregation).
Hence, a part-whole hierarchy can be viewed in terms of a number of general ``is-part-of’’ relations that can be reused between objects (see also \cref{sec:comp:relations}).

\subsubsection{Multi-Domain}
\label{sec:seg:multi_domain}

It is worth emphasizing that objects (as referred to in the context of this paper) are not restricted to vision, but also span sensory information from other domains such as audio or tactile\footnote{It is even discussed whether humans are capable of object perception in the olfactory domain~\citep{batty2014olfactory}.} (and even be entirely abstract, although this is not the focus of segregation).
For example, auditory objects may correspond to different sources of sound, such as speakers talking simultaneously in the same room (cocktail-party problem;~\citealp{cherry1953experiments}).
Objects in the tactile domain are perhaps less obvious, but consider the example of writing on a piece of paper with a pen, where you can clearly separate the sensations that arise from your fingers touching each other, touching the pen, and touching the paper~\citep[see also][]{kappers2015tactile}).
Notice how you are likely to associate the sensations of touching the pen and its visual perception with a common cause and therefore with the same object.
This implies that objects can be simultaneously grounded in sensory information from multiple domains, which may help resolve ambiguities (\eg McGurk Effect;~\citealp{mcgurk1976hearing}).

\subsection{Segregation Dynamics}
\label{sec:seg:dynamics}

Segregation needs not only infer a decomposition into objects, but also corresponding object representations. 
As is evident from our previous discussion, there is no universal choice of objects that is appropriate in all circumstances, which requires segregation to consider both context- and task-dependent information. 
Together with the need for a stable outcome, this has several consequences for the segregation dynamics which we will consider next.

\subsubsection{Multistability}
\label{sec:seg:multistability}

\begin{figure}
  \centering
  \includegraphics[width=\textwidth]{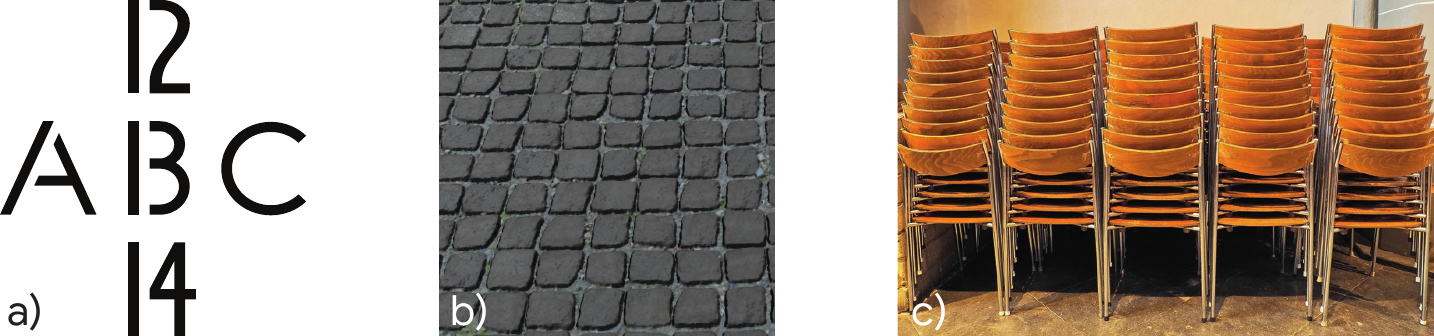}
  \caption{Human perception is multistable, which is often demonstrated using visual illusions as in \textbf{(a)}, yet it is also often encountered in the real world, \eg for different groupings of tiles \textbf{(b)}. To steer segregation towards a useful decomposition it is important to incorporate contextual information, for example to decide between a decomposition based on chairs or based on stacks in \textbf{(c)}.}
  \label{fig:seg:illusions}
\end{figure}

Most scenes afford many different useful decompositions that either stem from choosing different levels of granularity (\ie levels of hierarchy) or from ambiguous inputs that allow for multiple distinct but coherent interpretations (see multi-modal separation uncertainty \cref{sec:rep:reliability}).
Together, these result in a massive number of potential object representations (\eg $\geq 3000$ letters per page of text). 
Simultaneously representing all of them is not only intractable, but also undesirable, as the majority of object representations will not be useful for any particular situation. 
A practical solution to this problem is a dynamical segregation process that has \emph{multiple stable} equilibria that each correspond to a particular decomposition of a given scene.
Indeed, humans resolve this problem via multistable perception, which allows us to seamlessly switch back and forth between different interpretations~\citep{attneave1971multistability}.
This effect is often demonstrated with visual illusions as in \cref{fig:seg:illusions}a, but is in fact much more common than these constructed examples suggest.
For example, a simple tile pattern (as in  \cref{fig:seg:illusions}b) can easily be perceived in several ways, including rows or columns of tiles. 
Multistability can also be observed in other sensory modalities such as audio, tactile, and even olfaction~\citep{schwartz2012multistability}.
Notice that it is possible to simultaneously perceive multiple objects from the same decomposition, but not from different decompositions (\eg perceiving \texttt{13} and \texttt{B} simultaneously in \cref{fig:seg:illusions}a).
This inherent limitation of multistable segregation can also act as an advantage, since it ensures a single coherent decomposition of the input and avoids mixing objects from different incompatible decompositions. 
It implies that the process of segregation also has to be able to efficiently resolve conflicts from competing decompositions (explaining away).

\subsubsection{Incorporating Top-Down Feedback} 
\label{sec:seg:top_down}

Certain decompositions lead to a set of building blocks (objects) that are more useful than others for a given task or situation. 
For example, when moving a stack of chairs to another room it is useful to group information about the individual chairs together as a single object (see \cref{fig:seg:illusions}c).
On the other hand, when the goal is to count each of the individual chairs, a more fine-grained decomposition is preferred (and perhaps when repairing a chair an even more fine-grained decomposition is needed). 
These building blocks underlie the structure of downstream models that can be used for inference, prediction, and behavior, and the choice of decomposition therefore affects the ability to generalize in predictable and systematic ways.
Hence it is important that the outcome of the segregation process can be steered towards the most useful decomposition, based on contextual information.
One of the main sources of contextual information is \emph{top-down feedback}, for example in the form of task-specific information (\eg to guide visual search) or based on a measure of success at performing the given task.
Memory could act as another source of contextual information, for example by recalling a decomposition that has previously proven useful in the given situation.

\subsubsection{Consistency}
\label{sec:seg:consistency}

It is important that the grounding of object representations, as provided by the segregation process, is both stable and consistent across time (\ie it maintains object identity).
This helps to correctly accumulate partial information about objects, to infer temporal attributes from prior observations (\cref{sec:rep:temporal}), and to ensure that the outcome of more abstract computations in terms of object representations remain valid in the environment (\cref{sec:rep:reliability}).
It may also help to avoid ``double-counting'' of evidence (\eg during learning)\footnote{
Consider the example from~\citet{marcus2003algebraic} about owning a three-legged dog.
Despite the fact that you will likely see your dog much more often than other dogs, this series of observations does not affect your overall belief about the number of legs that dogs typically have, since these observations are all associated with \emph{the same} dog.}.
Object identity depends on a reliable mechanism for re-identification \ie a mechanism for identifying an object as being the same despite changes in appearance, perspective, or temporary absence of sensory information.
Consider, for example, a game of cups and balls, which involves tracking a ball hidden under one of three identical cups that are being moved around. 
In this case, a stable object identity requires maintaining separate identities for the cups despite their identical appearance, as well as re-identifying the ball as it reappears from under the cup.
When an object is re-encountered after a prolonged period, re-identification may require interfacing with some form of long-term memory.

\subsection{Methods}
\label{sec:seg:methods}

To succeed at segregation (in the sense outlined above) a neural network must acquire a comprehensive notion of objects and incorporate mechanisms to dynamically route their information.
Due to the prohibitive amount of potentially useful objects, it is unlikely that an adequate notion can be engineered directly or taught purely through large-scale supervision.
Therefore, in the following, we will review a wide range of approaches, including more classic non-neural approaches that have produced promising results despite incorporating domain-specific knowledge only to a lesser degree.
By also discussing the latter, we aim to provide inspiration for the development of neural approaches that can learn about objects directly from raw data (\eg by focusing on modularity).

\subsubsection{Clustering Approaches to Image Segmentation}
\label{sec:seg:methods:clustering}

\begin{figure}
  \centering
  \includegraphics[width=\textwidth]{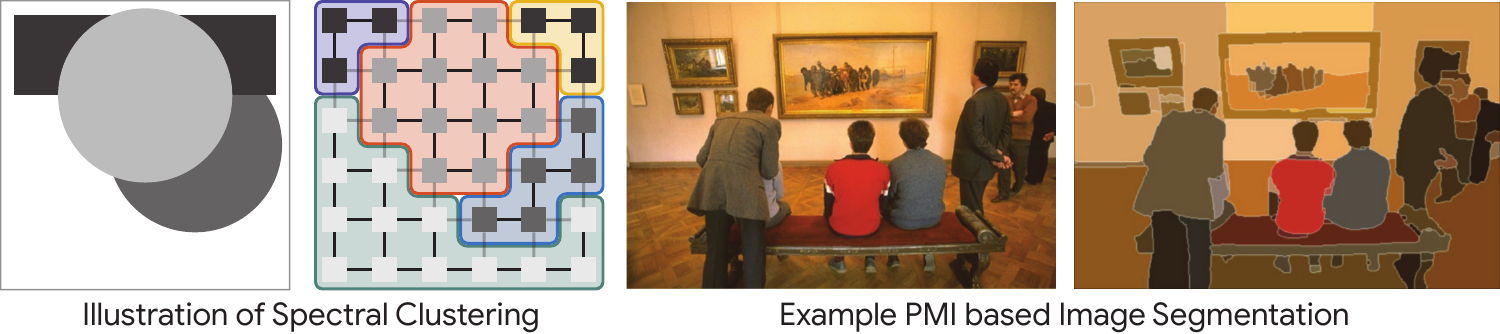}
  \caption{\textbf{Left:} An illustration of (spectral) clustering approaches, which treat image segmentation as a graph-partitioning problem. \textbf{Right:} Corresponding instance segments as obtained by \citet{isola2014crisp}.}
  \label{fig:seg:clustering}
\end{figure}

Image segmentation is concerned with segmenting the pixels~\citep[or edges][]{arbelaez2011contour} belonging to an image into groups (\eg objects) and therefore provides a good starting point for segregation.
A common approach to image segmentation is to cluster the pixels of an image based on some similarity function~\citep{jain1988algorithms}.
One particularly successful approach is the spectral graph-theoretic framework of normalized cuts~\citep{shi2000normalized}, which treats image segmentation as a graph-partitioning problem in which nodes are given by pixels and weighted edges reflect the similarity between pairs of (neighboring) pixels.
Partitioning is performed by trading-off the total dissimilarity between different groups with the total similarity within the groups.
To the extent that the similarity function is able to capture the predictive structure of the data, this is then analogous to the trade-off inherent to modularity. 
It is straightforward to achieve a hierarchical segmentation in this graph clustering framework, either via repeated top-down partitioning~\citep{shi2000normalized} or bottom-up agglomerative merging~\citep{mobahi2011segmentation,hoiem2011recovering}.

In the context of segregation, a central challenge is to define a good similarity function between pixels that leads to useful objects.
As we have argued, a hardwired similarity function~\citep[\eg as in][]{shi2000normalized,malik2001contour} has little chance at facilitating the required flexibility, although different initial seedings of the clustering may still account for multiple different groupings (\ie multistability).
Labeled examples can be used to address this challenge in a multitude of ways, \eg to learn a similarity function between segments~\citep{ren2003learning,endres2010category,kong2018recurrent} or discrete graphical patterns~\citep{lun2017learning}, to learn boundary detection~\citep{martin2004learning,hoiem2011recovering}, or as a means of top-down feedback~\citep{mobahi2011segmentation}.
Unsupervised approaches (based on self-supervision) provide a more promising alternative.
One approach is to learn a similarity function between pairs of pixels, \eg based on their point-wise mutual information using kernel-density estimation~\citep{isola2014crisp} or based on self-supervised prediction using a neural network~\citep{isola2015learning}.
Alternatively, one can attempt to steer the clustering process based on the unsupervised principle of compressibility (minimum description length;~\citealp{mobahi2011segmentation}).

Notice that, since clustering-based approaches to image segmentation focus on low-level similarity structures, their understanding of objects at a more high-level is limited (\ie at the level of object representations, but see \citealp{bear2020learning}).

\subsubsection{Neural Approaches to Image Segmentation}
\label{sec:seg:methods:Neural}
\begin{figure}
  \centering
  \includegraphics[width=\textwidth]{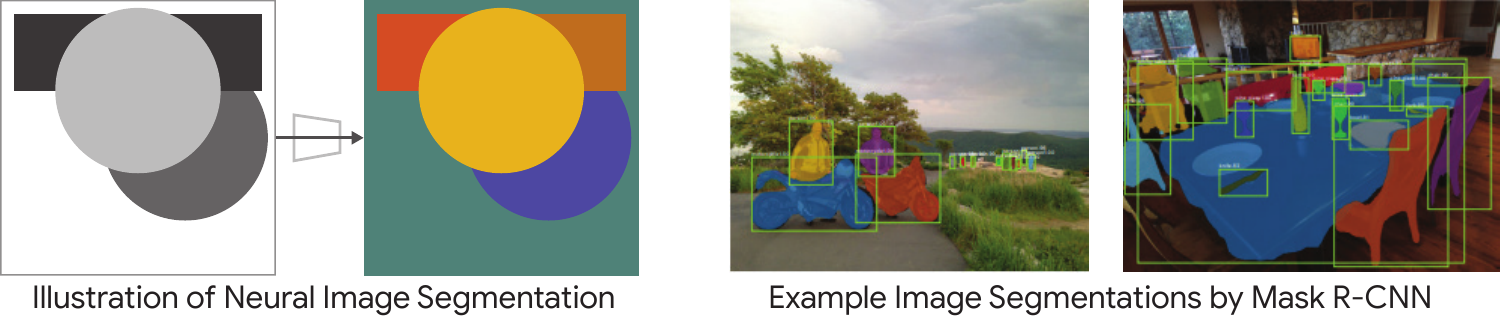}
  \caption{\textbf{Left:} An illustration of neural approaches that learn to directly output an image segmentation. \textbf{Right:} Corresponding bounding boxes and instance segments as obtained by \citet{he2017mask}.}
  \label{fig:seg:neural_segmentation}
\end{figure}

An alternative approach to image segmentation that leverages the success of end-to-end learning, is to directly output the segmentation with a deep neural network.
Unlike clustering-based approaches, which focus on the similarity structure between pixels (or small segments), learning now takes place at the (global) image level, which allows objects to be modeled at multiple levels of abstraction.
On the other hand, due to the one-to-one (feedforward) mapping from image to segmentation, it may now be more difficult to provide multiple different segmentations (multistability) or a hierarchical segmentation, for a given input.

Recent approaches based on supervised learning from ground-truth segmentation have produced high-quality instance segmentations of real-world images\footnote{
We would like to emphasize the distinction between \emph{instance} segmentation and \emph{semantic} segmentation. 
In the context of segregation we are more interested in the former, which is concerned with the more general notion of each segment being an object (instance).
In contrast, semantic segmentation associates a particular semantic interpretation (in the form of a label) with each segment, and therefore can not segregate multiple objects belonging to the same class.}.
For example, approaches based on R-CNN~\citep{girshick2014rich} decompose the instance segmentation problem into the discovery of bounding boxes using region-proposal networks~\citep{ren2015faster} and mask prediction~\citep{dai2016instanceaware,he2017mask} to provide instance segmentations.
The more recent DEtection TRansformer (DETR; \citealp{carion2020endtoend}) was able to integrate these stages into a single Transformer-based network using a bipartite matching loss.
Other approaches output an energy function from which the segmentation is easily derived, \eg based on the Watershed transformation~\citep{bai2017deep}. 
Instance segmentation has also been phrased as an image-to-image translation problem using conditional generative adversarial networks~\citep{mo2018instagan}.
Approximate instance segments can also be obtained as a by-product of performing some other task, such as learning to interpolate between multiple images~\citep{arandjelovic2019copy} or minimizing mutual information between image segments~\citep{yang2020learning}.

Unsupervised approaches that directly infer the segmentation (and that do not require large-scale supervision) are more relevant in the context of segregation, but have received far less attention. 
\Citep{ji2018invariant} propose to train a neural network to directly output the segment that an input belongs to by maximizing the mutual information between paired inputs in representational space (although it operates at the level of patches as opposed to the global image).
In the context of video, motion segmentation often produces segments that correspond to instances (provided that they move, \eg~\citealp{cucchiara2003detecting}), which can for example be learned through unsupervised multi-task learning~\citep{ranjan2019competitive}.

\subsubsection{Sequential Attention}
\label{sec:seg:methods:attention}
\begin{figure}
  \centering
  \includegraphics[width=\textwidth]{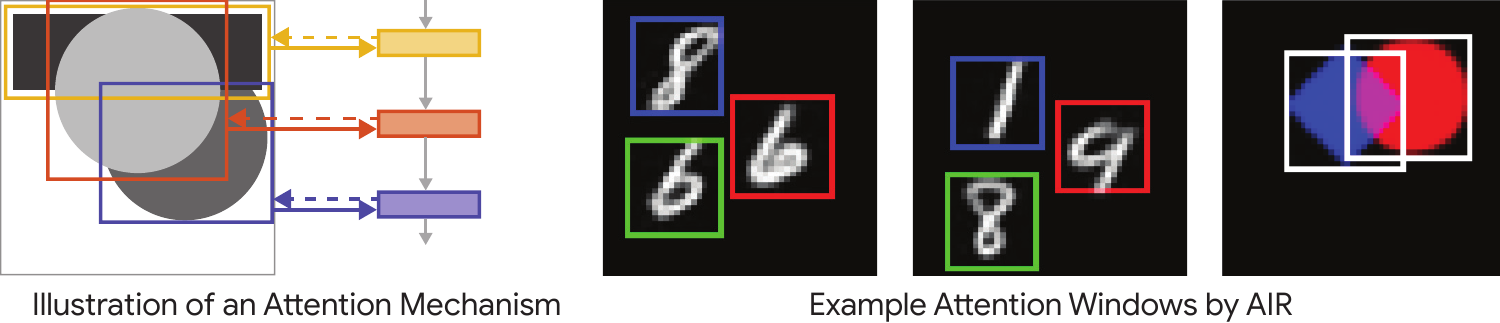}
  \caption{\textbf{Left:} An illustration of attention-based approaches, which sequentially attend to individual objects. \textbf{Right:} Corresponding attention windows as obtained by \citet{eslami2016attend}.}
  \label{fig:seg:attention}
\end{figure}

In the context of segregation, attention mechanisms provide a means to selectively attend to different objects sequentially.
Compared to image segmentation, this does not require exhaustively partitioning the image but instead allows one to focus only on the relevant locations in the image (\eg as a result of top-down feedback).
Here we focus mainly on \emph{hard} attention mechanisms that attend to a strict (\ie spatially delineated) subset of the available information in the form of an attention window, \eg in the shape of a bounding-box~\citep{stanley2004evolving} or a fovea~\citep{schmidhuber1991learning}.
Their strong spatial bias (due to the shape of the attention window) makes them particularly relevant for the domain of images, but more difficult to adapt to modalities in which meaningful objects are not characterized by spatial closeness.
On the other hand, the rigid shape of the attention window may interfere with modularity due to potential difficulties in extracting information about objects with incompatible shapes or that are subject to occlusion. 

The main challenge for incorporating attention mechanisms is in correctly placing the window.
Early approaches by-pass this problem by evaluating a fixed attention window exhaustively at each possible image location, or using several of many heuristics~\citep{lampert2008sliding,alexe2010what,uijlings2013selective}.
A classifier can then be trained to determine which window contains an object~\citep{rowley1998neural,viola2001rapid,harzallah2009combining}.
Other approaches compute a two-dimensional topographical saliency map that reflects the presence of perceptually meaningful structures at a given location.
This facilitates an efficient control strategy to direct an attention window in an image by visiting image locations in order of decreasing saliency~\citep{itti1998model}.
Salient regions can be learned based on bottom-up information, such as the self-information of local image patches~\citep{bruce2006saliency}.
Alternatively, they can be derived by also incorporating top-down information, \eg by highlighting locations that are (maximally) informative with respect to a discriminative task~\citep{gao2005discriminant,cao2015look,zhmoginov2019informationbottleneck}.  
Recently, there has been renewed interest in saliency-based approaches through the discovery of keypoints~\citep{jakab2018unsupervised, kulkarni2019unsupervised, minderer2019unsupervised,gopalakrishnan2020unsupervised}.

It is also possible to directly learn the control strategy for placing the window of attention, which naturally accommodates top-down feedback.
For example, learning the control strategy can be viewed as a reinforcement learning problem, in which the actions of an ``agent'' determine the location of the window.
A policy for the agent (frequently implemented by a neural network) can then be evolved~\citep{stanley2004evolving}, trained with Q-learning~\citep{paletta2005qlearning}, or via Policy Gradients~\citep{butko2009optimal}.
Alternatively, it can be incorporated as a separate action in an agent trained to perform some task (\eg classification) or to interact with an environment~\citep{mnih2014recurrent,ba2014multiple}.
AIR~\citep{eslami2016attend} and its sequential extension SQAIR~\citep{kosiorek2018sequential} deploy a similar strategy for an unsupervised learning task with the purpose of extracting object representations.
They make use of an attention mechanism that is fully differentiable based on spatial transformer networks~\citep{jaderberg2015spatial}, but see also DRAW~\citep{gregor2015draw} for an alternative mechanism.
Similarly, \cite{tang2014learning} incorporates a window of attention in a deep belief network to extract object representations by performing (stochastic) inference over the window parameters alongside the belief states.

\emph{Soft} attention mechanisms implement attention as a continuous weighing of the input (\ie a mask) and can be seen as a generalization of hard attention.
For example, in MONet~\citep{burgess2019monet}, GENESIS~\citep{engelcke2019genesis}, and ECON~\citep{vonkugelgen2020causal} a recurrent neural network is trained to directly support the learning of object representations by outputting a mask that focuses on different objects at each step\footnote{
Notice, however, that these particular methods enforce an \emph{exhaustive} partition of the image similar to image segmentation methods.
}.
A similar soft-attention mechanism has also been used to facilitate supervised learning tasks such as caption generation~\citep{xu2015show}, instance segmentation~\citep{ren2017endtoend}, or (multi-)object tracking~\citep{kosiorek2017hierarchical,fuchs2019endtoend}.
Soft attention mechanisms have also been applied \emph{internally} (self-attention) to support segregation.
For example, \citet{mott2019interpretable} incorporates a form of dot-product attention~\citep{vaswani2017attention} in an agent to attend to the internal feature maps of a bottom-up convolutional neural network that processes the input image.
A similar self-attention mechanism was also used to support image classification~\citep{zoran2019robust}.

\subsubsection{Probabilistic Generative Approaches}
\label{sec:seg:methods:generative}
\begin{figure}
  \centering
  \includegraphics[width=\textwidth]{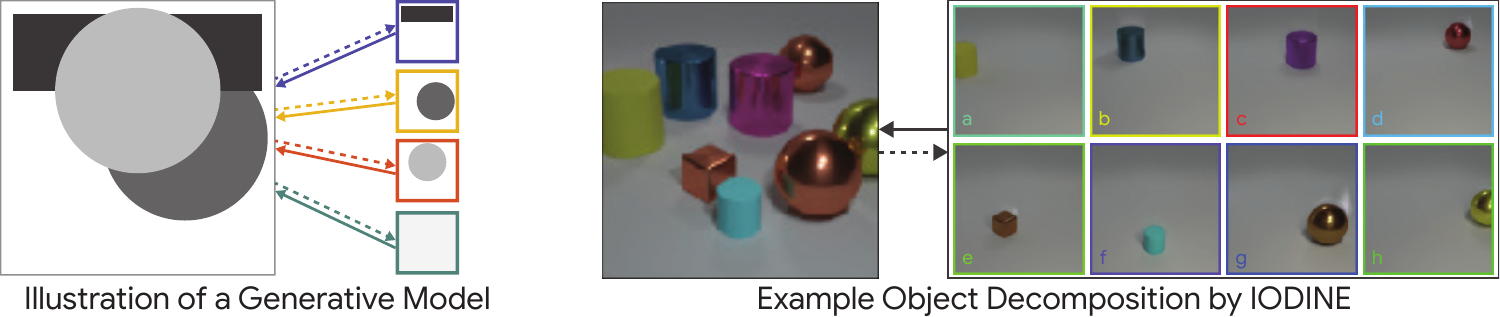}
  \caption{\textbf{Left:} An illustration of generative approaches to segregation that model an image as a mixture of components. \textbf{Right:} A corresponding decomposition in terms of individual objects as obtained by \citet{greff2019multiobject}.}
  \label{fig:seg:generative}
\end{figure}

A probabilistic approach to segregation is via inference in a generative model that models the observed data in terms of multiple components (objects)
\footnote{
Human perception is also said to be generative in the sense that we often perceive objects as coherent wholes even when they are only partially observed (amodal completion;~\citealp{michotte1991amodal}). 
}.
An advantage of explicitly modeling the constituent objects is that it is easy to incorporate assumptions about their structure, including modularity and hierarchy.
This then enables inference (segregation) to go beyond low-level similarities or spatial proximity, and recover object representation based on their high-level structure as implied by the model.
On the other hand, as we will see below, inference usually becomes more difficult as the complexity of the generative model increases, and especially when considering multi-modal distributions (\ie for multistability).

The most basic assumption to incorporate in a generative model, for the purposes of segregation, is to assume that the input is \emph{directly} composed of multiple parts (objects) that are each modeled individually.
Inference in such models then allows one to recover a partitioning of the input in addition to a description of each part (object representation).
Early approaches model images with a \emph{mixture model} that treats the color values of individual pixels as independent data points that are identically distributed~\citep{samadani1995finite,friedman1997image}.
Alternatively, the decomposition can be based on other features such as optical flow~\citep{jepson1993mixture} or the coefficients of a wavelet transform~\citep{guerrero-colon2008image}.
Mixture models can also be biased towards spatial coherence to explicitly account for the spatial structure of visual objects~\citep{weiss1996unified,blekas2005spatially}.
Independent Component Analysis (ICA) models the observed data as linear combinations (mixtures) of unobserved random variables (sources) that are statistically independent~\citep{hyvarinen2000independent}. 
This approach has been particularly successful at blind source separation (segregation) in the auditory domain (\eg the cocktail party problem~\citealp{cherry1953experiments}), although it has also seen application in the context of images~\citep{lee2002unsupervised}.

To more accurately model complex data distributions, it is possible to incorporate domain-specific assumptions in the generative model (and thereby improve the result of inference).
For example, a generative model that captures the geometry of 3D images of indoor scenes as well as the objects that are in it ``[\dots] integrates a camera model, an enclosing
room `box', frames (windows, doors, pictures), and objects (beds, tables, couches, cabinets), each with their own prior on size, relative dimensions, and locations''~\citep{delpero2012bayesian}.
The results that can be obtained by incorporating domain-specific knowledge are impressive~\citep{zhao2011image, delpero2012bayesian, delpero2013understanding, tu2005image, tu2002image}.
However, performing inference in highly complex generative models of this type is problematic and frequently relies on custom inference methods tailored to this particular task (\eg Markov Chain Monte Carlo using jump moves to remove or add objects or specific initialization strategies).
In recent years, \emph{probabilistic programming languages} have emerged as a general-purpose framework to simplify the design of complex generative models and the corresponding inference process.
For example, they have enabled the use of symbolic graphic renderers as forward models~\citep{mansinghka2013approximate} and incorporated deep neural networks to help make inference more tractable~\citep{kulkarni2015picture,romaszko2017visionasinversegraphics}.
Nonetheless, in the context of segregation, the amount of domain-specific engineering that is still required limits their generality and applicability to other domains (similar to overly relying on supervised labels from a particular domain).

An alternative approach to more accurately modeling complex data distributions is to incorporate fewer assumptions, and rather parameterize the generative model with a neural network that can \emph{learn} a suitable generative process from many different observations.
For example, \cite{vansteenkiste2019investigating} demonstrates how a (spatial) mixture model that combines the output of multiple deep neural networks is able to learn to generate images as compositions of individual objects and a background~\citep[see also][]{nguyen-phuoc2020blockgan,ehrhardt2020relate,niemeyer2020giraffe}.
However, in order to perform segregation, we must also be able to perform inference in these models, which can be very challenging.
This has been addressed by simultaneously learning an amortized iterative inference process based on de-noising~\citep{greff2016tagger}, generalized expectation-maximization~\citep{greff2017neural}, iterative variational inference~\citep{greff2019multiobject}, slot attention~\citep{locatello2020objectcentric}, or parallel spatial (bounding-box) attention~\citep{lin2020space,jiang2020generative}.
Further improvements can be made by assuming access to multiple different views to explicitly model 3D structure at a representational level~\citep{chen2020objectcentric,nanbo2020learning}.
Even though these methods still struggle at modeling complex real-world images, they are capable of learning object representations that incorporate many of the previously mentioned desiderata (\eg common format, disentangled, modular), in a completely unsupervised manner.

\subsection{Learning and Evaluation}
\label{sec:seg:summary}

The main challenge in segregation is in coping with the immense variability of useful objects that depend on both task and context.
We have argued that this effectively precludes solutions that overly rely on supervision or domain-specific engineering.
This raises the question of how a useful notion of an object can be discovered mainly via unsupervised learning (and later refined based on task-specific information).
A key part of the answer is to focus on the modularity of objects, which only depends on the statistical structure of the observed data and interfaces directly with the functional role of objects as compositional building blocks.
Indeed, evidence suggests that human object perception is based on similar principles~\citep{orban2008bayesian,chater1996reconciling}.
In the machine learning literature, several approaches have also shown to be able to successfully leverage modularity to learn about objects, either in combination with spectral clustering~\citep{isola2014crisp}, attention~\citep{burgess2019monet}, or by using neural mixture models~\citep{greff2019multiobject}, or an adversarial formulation~\citep{yang2020learning}.
Additionally, also focusing on other properties of objects such as common fate (\eg motion) may play an important role in further improving these results~\citep[\eg][]{pathak2016learning,ranjan2019competitive}.

Regarding segregation dynamics, we have seen that it is important to provide architectural inductive biases that help with dynamic information routing, \eg in the form of attention or masking specific parts of the input. 
Consistency and top-down feedback are mostly affected by the interplay between segregation, representation, and composition, and it is difficult to evaluate these properties in isolation. 
However, in order to facilitate this interaction, it is critical that segregation is part of a fully-differentiable neural approach, which may be most problematic for clustering-based approaches to image segmentation and probabilistic programs based on symbolic models.

Segregation is best evaluated in the context of a larger system, where the resulting object representations form the foundation of structured models for inference, behavior, and prediction. 
In this case, the ability to transfer learned object representations to other tasks, and improving sample-efficiency (semi-supervised) is of particular interest~\citep{wei2019fss1000}.
Alternatively, when ground-truth information about objects is available, individual aspects of segregation can be evaluated more directly.
For example, when a pixel-level segmentation is produced as part of segregation, then metrics such as AMI~\citep{vinh2010information} can be used to compare against the ground-truth.
This also provides a means to probe multi-stability for inputs that are known to have multiple stable interpretations.
Finally, consistency can also be evaluated in this way, namely by measuring how stable the inferred notion of an object is across a temporal sequence (\eg object tracking).
}
{
\section{Composition}
\label{sec:comp}

\newcommand{\objA}{$\textcolor{viridis_green}\blacksquare$\xspace}
\newcommand{\objB}{\textcolor{viridis_blue}\textbullet\xspace}
\newcommand{\objC}{$\textcolor{viridis_petrol}\bigstar$\xspace}

\begin{figure}
    \includegraphics[width=\textwidth]{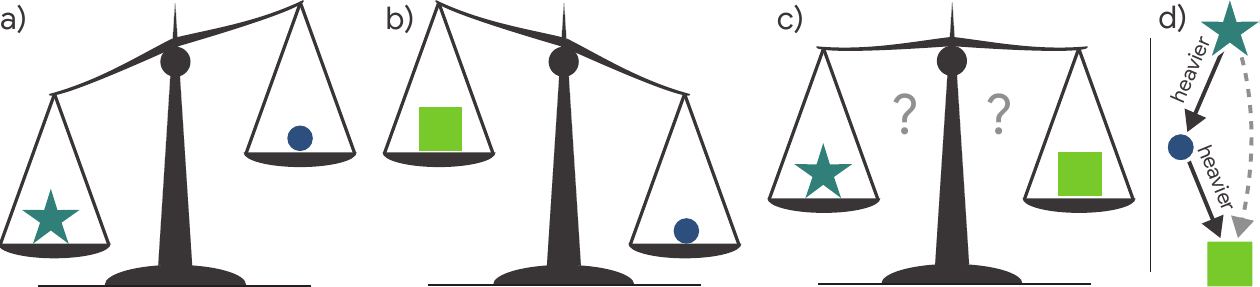}
    \caption{Three different objects (\objA, \objB, \objC) appear in different pairings on a
scale \textbf{(a)} and \textbf{(b)}.
   By evaluating their relationships \textbf{(d)}, it can be inferred how the
scale will tip in \textbf{(c)}.}
    \label{fig:comp:entailment}
\end{figure}

In this section, we look at the binding problem from the perspective of composition: building structured models of the world that are \emph{compositional}.
Here we encounter the need for variable binding: the ability to combine object representations and relations without losing their integrity as constituents (as is needed for compositionality).
As we have seen in \cref{sec:bind}, compositionality is a core aspect of human cognition and underlies our ability to understand novel situations in terms of existing knowledge.
Similarly, in the context of AI, it supports the systematic reuse of familiar objects and relations to dynamically construct novel inferences, predictions, and behaviors, as well as the ability to efficiently acquire new concepts in relation to existing knowledge.

Consider the sequence of observations in \cref{fig:comp:entailment}, which allows you to infer the relative weights of the three depicted objects (\objA, \objB and \objC).
Several interesting observations can be made.
For example, from panel (a) you can tell that \objB is heavier than \objA, and likewise, that \objC is heavier than \objB from panel (b).
This information does not describe a property of any of the individual objects, but rather a \emph{relation} between them.
On the other hand, it can still be used to update the properties of the participating objects in response to new information (\eg the precise weight of \objA) or to respond to generic queries, such as answering which of the objects is the heaviest.
The latter, in this case, also requires comparing the weights of \objA and \objC (panel (c)).
Notice how this is only possible through transitivity of the “heavier than” relation, which allows you to combine the relations from panels (a) and (b) to infer that \objC is heavier than \objA. 

In the following, we take a closer look at how to enable neural networks to dynamically implement \emph{structured models} for a given task, with the ultimate goal of generalizing in a more systematic (human-like) fashion.
First, we focus on incorporating a compositional structure that combines relations and object representations without undermining their modularity (\cref{sec:comp:structure}).
Next, we consider how a neural network can dynamically infer the appropriate structure and leverage it for the purpose of reasoning (\cref{sec:comp:reasoning}).
Towards the end, we survey relevant approaches from the literature that address these aspects of composition (\cref{sec:comp:methods}).

\subsection{Structure}
\label{sec:comp:structure}


To implement structured models, a neural network must organize its computations to reflect the desired \emph{structure} in terms of objects and their relations.
This structure is generally described by a graph where nodes correspond to objects and edges to relations\footnote{
In our discussion, we focus mainly on binary relations (\eg A is bigger than B) that are well represented by individual edges.
However, keep in mind that it is also possible to represent higher-order relations (\eg A divides B from C), either by using a higher-order graph (\eg a factor graph) or with the help of auxiliary nodes (\eg by adding a `division node’ with binary relations to A, B, and C).
}.
By representing relations separately (independent of object representations) it is possible to freely compose relations and objects to form arbitrary structures (\ie corresponding to different graphs).
However, certain types of relations may also impose constraints on the structure to ensure internal consistency between relations (\eg symmetry, transitivity).

\begin{SCfigure}[][t]
    \includegraphics[width=0.6\textwidth]{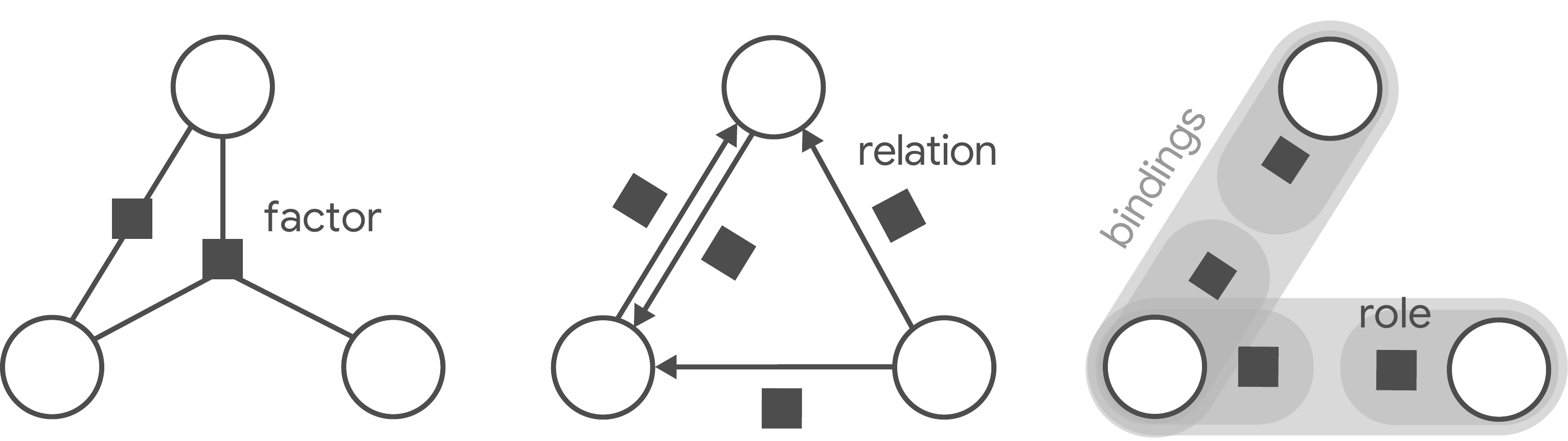}
    \caption{Three different ways in which structure can be defined in terms of
relations between objects: As a factor graph, a directed graph, or as nested role-filler bindings.}
    \label{fig:graph-fig-doc}
\end{SCfigure}

\subsubsection{Relations}
\label{sec:comp:relations}

Relations encode the different computational interactions between the object representations in a structured model.
Many different types of relations are possible, including causal relations (\eg ``collides with''), hierarchical relations (``is part of''), or comparative relations (\eg ``bigger than'').
Moreover, these general relations can often be specialized to include the nature or strength of an interaction (\eg ``\emph{elastic} collision'', ``\emph{much} bigger than’’).
To efficiently account for this variability and support learning, relations are best encoded using flexible (neural) representations.
Similar to object representations, it may then also be desirable to use a common format that provides a measure of similarity between relations and ensures that they can be used interchangeably\footnote{
\Citet{doumas2008theory} even argues that objects and relations should in fact use a  \emph{shared} `feature pool' with which both can be described.
}.
The way structure is defined in terms of relations may also have implications for their corresponding representations.
When the structure is given by a regular (directed) graph or a factor graph (see \cref{fig:graph-fig-doc} a \& b), then each relation is encoded by a single representation corresponding to either an edge or a factor.
Alternatively, it is possible to encode a relation with multiple representations that correspond to the different \emph{roles} that the participating objects play (see \cref{fig:graph-fig-doc} c).
Finally, it is important that relations are represented separate from and independent of the object representations (see also \emph{role-filler-independence};~\citealp{hummel2004solution}).
This enables relations and objects to be composed in arbitrary ways to form a wide variety of (potentially novel) structures.

\subsubsection{Variable Binding}
\label{sec:comp:variable_binding}

To enable a single neural network to implement different structured models, it requires a suitable `variable binding’ mechanism\footnote{
The term variable binding is adapted from mathematics, where it refers to binding the free variables in an expression to specific values.
In our case, variables correspond to object representations that are bound to the structure determined by the relations.
} that can \emph{dynamically} combine modular object representations and relations.
Consider the classic example of Mary and John adapted from \citet{fodor1988connectionism}:
Depending on a given task or context it may be more important to consider that “Mary loves John”, that “John is taller than Mary”, or that “Mary hit John”.
In general, the number of possible structures that can be considered is potentially very large, and it is, therefore, intractable to represent all of them simultaneously.
Apart from being dynamic, a suitable variable binding mechanism should also preserve the modularity of individual object representations.
This is critical to implement structured models that are \emph{compositional}, which ensures that the neural network generalizes systematically and predictably with respect to the underlying objects.

In many cases, only a single level of variable binding that directly combines individual object representations and relations is needed. 
However, in certain other cases (\eg “Bob knows that Mary loves John”) it may be required to first build composite structures that can themselves act as `objects’, and that can then be combined recursively.
When using a role-based representation for relations, multiple levels of variable binding are also needed to avoid ambiguity when a low-level object representation plays the same role in multiple relations.\looseness=-1

\subsubsection{Relational Frames}
\label{sec:comp:relational_frames}

\begin{figure}
   \includegraphics[width=\textwidth]{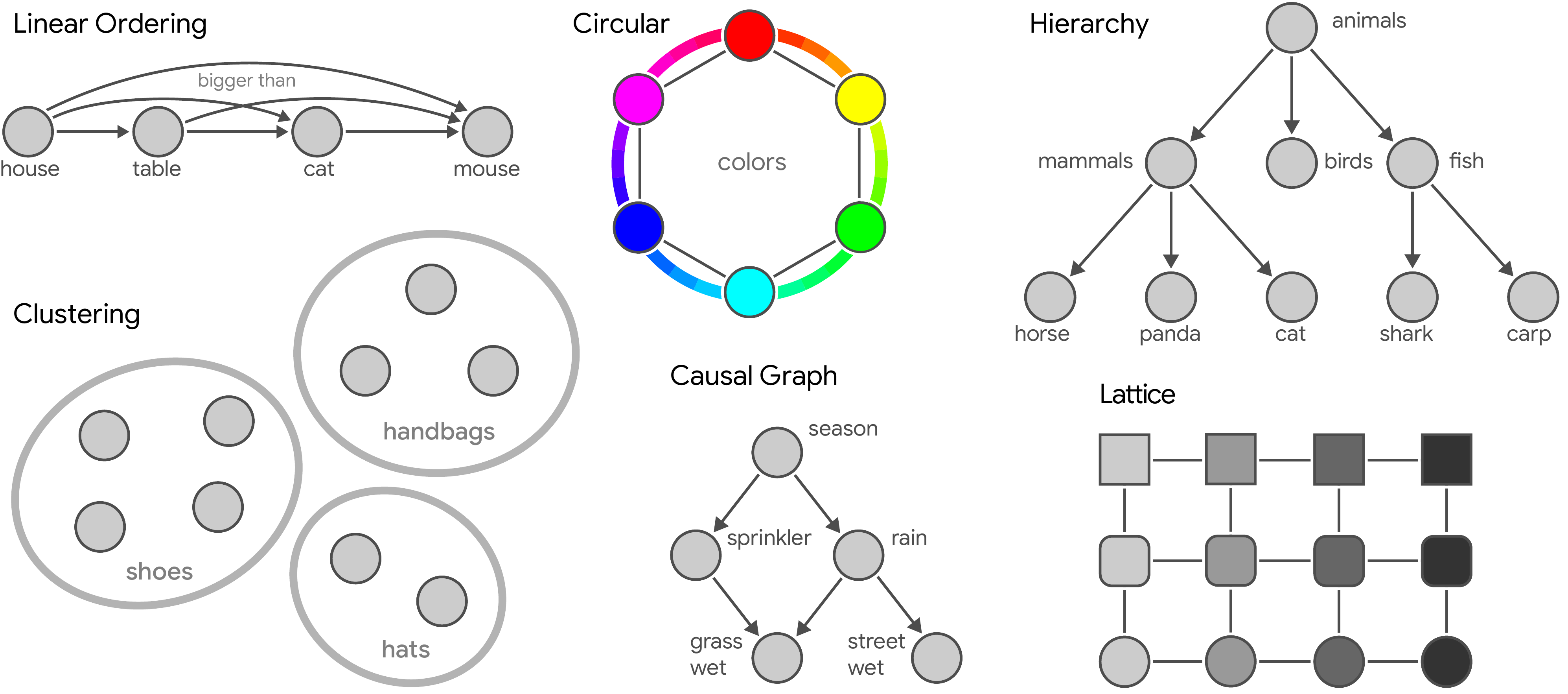}
   \caption{Examples of different structural forms \citep{kemp2008discovery} that each can be used to define relations among objects and imply different patterns of generalization.}
   \label{fig:comp:structure}
\end{figure}

Each \emph{type} of relation focuses on a particular aspect of the broader interaction among objects, and thereby defines a particular \emph{relational frame} that is internally consistent.
Consider again the example in \cref{fig:comp:entailment}, which was concerned with the “heavier than” relation. 
This corresponds to a relational frame of comparison that induces an ordering among the objects in terms of their weight.
In this case, an internally consistent ordering requires the relation to be transitive (\ie $A > B \cap B > C \Rightarrow A > C$) and anti-symmetric (\ie $A > B \Rightarrow B \not> A$).
More generally, a relational frame is characterized by a particular type of relation, and by the logical consequences (\ie different entailments) that are implied by having (multiple) relations of this type within the structure.
We adopted the term relational frame from Relational Frame Theory (RFT; see also \cref{sec:rel:rft}), which distinguishes two types of entailment that humans primarily use to derive (unobserved) relations: \emph{mutual entailment} and \emph{combinatorial entailment}.
Mutual entailment is used to derive additional relations between two objects based on a given relation between them, \eg anti-symmetry for a frame of comparison, or symmetry for a frame of coordination (\ie deriving $B = A$ from $A = B$).
Analogously, combinatorial entailment is used to derive new relations between two objects, based on their relations with a shared third object, \eg transitivity for a frame of coordination (\ie deriving $A=C$ from $A=B$ and $B=C$).

Many different types of relational frames can be distinguished, which can be organized into a number of general classes~\citep{hughes2016relational}, including `coordination' (\eg same as) , `comparison' (\eg larger than), `hierarchy' (\eg part of) , `temporal’ (\eg after), or `conditional' (\eg if then).
Their corresponding rules for entailment give rise to different \emph{structural forms}~\citep{kemp2008discovery} among their relations, such as trees, chains, rings, and cliques (see \cref{fig:comp:structure}).
In this way, each relational frame can also be seen as encoding a particular (systematic) \emph{pattern of generalization} among the objects.
Multiple different relational frames may co-occur within the same structure, which allows for rules of entailment to interact across different frames to facilitate more complex generalization patterns (\eg $A=B$ and $B>C$ implies $A>C$).

\subsection{Reasoning}
\label{sec:comp:reasoning}

The appropriate structure for a model depends on the task and context, and should therefore be dynamically inferred by the neural network to focus only on relevant interactions between the objects.
Likewise, it is important to consider the computational interactions between relations and object representations, in order to make use of the inferred structure for prediction and behavior.

\subsubsection{Relational Responding}
\label{sec:comp:relational_responding}

To leverage a given structure in terms of relations between object representations, a neural network must be able to organize its computations accordingly.  
A common use case involves adjusting the (task-specific) response to an object based on its relation to other objects (relational responding).
For example, if it is known that \objA is heavier than \objB, then learning that \objB is too heavy for a particular purpose (task) also changes your behavior concerning \objA.
More generally, relational responding of this kind may involve evaluating multiple (derived) relations between objects and combining information across different relational frames.
Another use case is in implementing so-called \emph{structure sensitive operations}~\citep{fodor1988connectionism} that require responding \emph{directly} to the structure given by the relations (independent of the object representations).
This is especially important for solving abstract reasoning tasks, \eg when applying the distributive law to a given mathematical expression (\ie turning $a \cdot (b + c)$ into $a \cdot b + a \cdot c$). 

A natural choice for facilitating relational responding in a neural network is to organize its internal information flow (\ie computations) in a way that reflects the graph structure of relations and objects.
This ensures that newly available information affects the object representations in accordance with the dependency structure implied by the relations (and therefore also with the generalization patterns due to the relational frames).
Most information processing of this kind can then be implemented in terms of only local interactions between objects representations and relations, which maximally leverages their modularity.
These local interactions, which can either be instantaneous (\eg collides with) or persistent (\eg is part of), can facilitate both directed (\eg for causal relations) and bidirectional (\eg for comparison) information flow.
On the other hand, local interactions are ill-suited for implementing structure sensitive operations that require simultaneously considering multiple different parts of the larger structure.

\begin{SCfigure}[][t]
  \centering
  \includegraphics[width=0.6\textwidth]{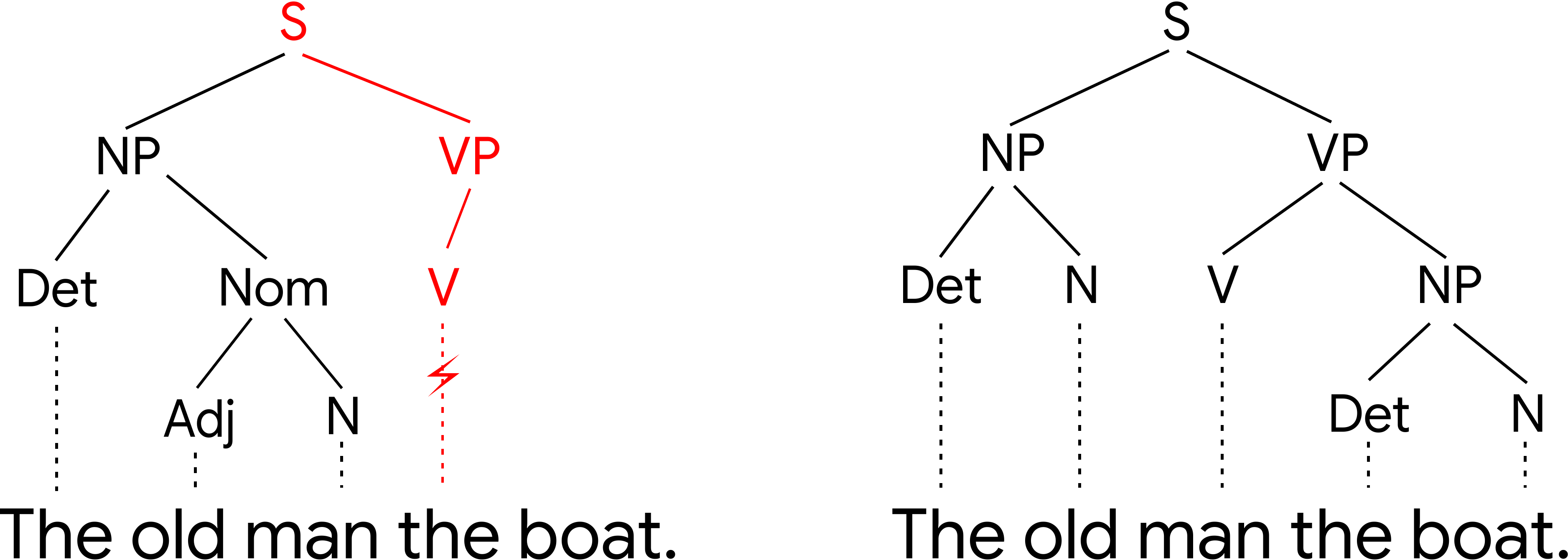}
  \caption{Two parse-trees of a garden-path sentence: The intuitive parsing (on the \textbf{left}) fails, even though the sentence is grammatically correct (see parse-tree on the \textbf{right}).}
  \label{fig:comp:the_old_man}
\end{SCfigure}

\subsubsection{Inferring Structure}
\label{sec:comp:inferring_structure}

Inferring the most desirable structure is an inherently difficult task, which requires making many individual choices at the level of relations that all have to be coordinated to ensure that the structure as a whole is useful.
One important guiding constraint is the internal consistency of the structure with respect to the rules of entailment as implied by the choice of relational frames.
Inconsistencies between the observed information and predictions by the structured model are another indicator of a wrong or incomplete structure.
The `garden-path’ sentence ``The old man the boat.'' (see \cref{fig:comp:the_old_man}) provides a good example for a violation of expectations, which then triggers a revision of the structure.
Upon first reading, ``The old man'' is likely parsed as the subject of the sentence, which implies a structure where the next word is expected to be a verb.
However, since ``the boat'' is not a verb (and therefore does not match this expectation), the sentence cannot be parsed in this way.
The problem is resolved by revising the structure so that it takes ``The old'' as the subject and ``man'' as the \emph{verb} of the sentence.
This example also illustrates the need for collaboration between composition and segregation:
It was the initial grouping of ``The old man'' as a single object that gave rise to inconsistencies at the level of structure, which could only be resolved by also changing the outcome of the segregation process.
Hence, it is vital that the process of inferring structure is able to provide (top-down) feedback to help guide the process of segregation. 

Inferring structure at the level of individual relations between objects involves making choices about the type of relation, or which of the properties of an object to relate.
These decisions can be guided by \emph{contextual cues} from the environment, such as the scales in \cref{fig:graph-fig-doc} that trigger a comparison of the objects in terms of their masses (as opposed to \eg their relative position or shape).
Inferring a relation between objects may also be triggered upon discovering their relation to other objects (\eg due to combinatorial entailment).
However, for the sake of efficiency it may not always be desirable to explicitly represent such relations, but rather model their effect implicitly due to appropriately organizing the computations of the network (\ie relational responding).
More generally, the process of inferring structure has to interface closely with the mechanism for variable binding (\ie for dynamically combining modular object representations and relations in a way that preserves their modularity).

\subsection{Methods}
\label{sec:comp:methods}

To succeed at composition, a neural network requires a mechanism for organizing its internal computations in a way that facilitates relational responding based on the desired structure.
A natural approach is to incorporate the structure at an \emph{architectural level} by focusing directly on the local interactions between objects representations and relations. 
Alternatively, one can also use a more generic (recurrent) neural network “processor” that (sequentially) operates on a \emph{representation} of the desired structure.
In the following we will review both of these different approaches, focusing in particular on relational responding and the difficulty of inferring structure\footnote{We note that the problem of inferring structure has also received considerable attention in the causality literature, often specifically focusing on cause-effect discovery (\eg see \citet{hoyer2009nonlinear,lopez-paz2015learning,peters2016causal} or \citet{peters2017elements} for an overview).
Generally, we expect structural causal models to become highly relevant for composition, due to their robustness under intervention and utility for reasoning about hypothetical or unobserved scenarios~\citep{pearl2019seven,scholkopf2019causality}.
}.

\subsubsection{Graph Neural Networks}
\label{sec:comp:graph_neural_networks}

Graph Neural Networks (GNNs;~\citealp{scarselli2009graph,pollack1990recursive}) are a promising approach for composition that incorporates the desired structure for relational responding at an architectural level (see \citealt{wu2019comprehensive} for an overview).
At a high level, a GNN is a neural network that is structured according to a graph whose edges determine how information is exchanged among the nodes.
In the context of composition, nodes correspond to object representations and edges to relations, which together form the structure, \ie using (static) variable binding at the architectural level.
A GNN fundamentally distinguishes two kinds of information processing, one that requires evaluating the relations between the object representations, and another that is concerned with combining (aggregating) the effect of the incoming relations to update the object representations~\citep{battaglia2018relational}.
By implementing these in a general way that applies equally to different objects and relations, a GNN can accommodate many different structures.
In general, the local information processing in a GNN ensures that information affects the object representations in a way that follows the dependency structure implied by the relations (relational responding).

\paragraph{Graph Convolutional Networks}
Graph Convolutional Networks (GCNs) are a type of GNNs based on a generalization of convolutional neural networks (which operate on grids) to non-Euclidean geometries such as graphs~\citep{bronstein2016geometric}.
A GCN consists of several layers that each produce an updated set of node representations by applying graph-convolutions to a local neighborhood in the graph.
They have been successfully applied to a wide variety of graph-structured data including social networks~\citep{hamilton2017inductive}, citation networks~\citep{kipf2016semisupervised}, 3D surfaces~\citep{litany2018deformable}, knowledge base completion tasks~\citep{schlichtkrull2018modeling}, and bio-chemical modeling~\citep{atwood2016diffusionconvolutional}.
However, while they excel at modeling large-scale graphs, one disadvantage of GCNs in the context of composition is that they assume a given graph in the form of an adjacency matrix and node representations as input.
For the purpose of composition, scalability is less important since we are most interested in relatively small graphs (restricted by working memory) that are composed dynamically.
On the other hand, some GCNs~\citep[\eg][]{henaff2015deep,lee2019selfattention} have used a mechanism for coarsening (down-sampling) the graph between layers, to reduce computational complexity, which could provide a mechanism for refining the structure (\ie structure inference).

\paragraph{Message Passing Neural Networks}
Message Passing Neural Networks (MPNNs; \citealp{gilmer2017neural}) iteratively update the node representations of a given graph by exchanging messages along its edges (until convergence)\footnote{
Recently, MPNNs were extended to allow for continuous updates~\citep{deng2019continuous,liu2019graph}.
}.
Compared to GCNs, both the graph structure and weights are shared across layers (iterations), and the messages (corresponding to the incoming relations) are typically implemented as a pairwise \emph{non-linear} function of both adjacent node representations.
Hence, edges play a more prominent role in information processing and by explicitly considering pair-wise interactions it is easier to model comparative relations between objects.
MPNNs were initially conceived as a generalization of RNNs to graph-structured inputs~\citep{sperduti1997supervised,gori2005new} and have since been adapted to consider modern deep neural networks~\citep{li2016gated}.
A more general framework that accommodates both MPNNs and GCNs was proposed in \cite{battaglia2018relational}, which additionally includes a global representation of the graph that interacts with all the nodes and edges (and may thereby more easily provide for structure-sensitive operations).

MPNNs have been shown to generalize more systematically (compared to standard neural networks) on many different tasks that require relational responding in terms of objects, including common-sense physical reasoning~\citep{chang2016compositional,battaglia2016interaction,janner2018reasoning}, hierarchical physical reasoning~\citep{mrowca2018flexible,li2020visual,stanic2020hierarchical}, visual question answering~\citep{santoro2017simple,palm2018recurrent}, abstract visual reasoning~\citep{andreas2019measuring}, natural language processing~\citep{tai2015improved}, physical construction~\citep{hamrick2018relational} or multi-agent interactions~\citep{sun2019stochastic}.
Similar to GCNs, the desired structure may either be specified directly or inferred dynamically based on some heuristic, \eg based on proximity~\citep{chang2016compositional,mrowca2018flexible} or a language parser~\citep{tai2015improved}.
Alternatively, MPNNs have been used to implement a relational inductive bias based on a generic structure, \eg by assuming it to be fixed and fully connected (as in Relation Networks; \citealp{santoro2017simple}).
In this case, information can still be exchanged among all the nodes, although the generalization implied by having the correct structural dependencies is lost (\eg for entailment).

A more desirable approach is to (dynamically) infer the desired structure, although this is challenging due to the discreteness of graphs and difficulties in comparing them efficiently. 
One approach is to first learn a continuous embedding for all possible graph structures and then optimize for the right structure in the corresponding space, \eg using VAEs~\citep{kusner2017grammar,zhang2019dvae}, or GANs~\citep{yang2019conditional}.
The other approach is to directly infer the connectivity between nodes iteratively based on message passing, \eg for a fixed number of nodes as in Neural Relational Inference (NRI; \citealp{kipf2018neural}) or adaptively as in Graph Recurrent Attention Networks (GRANs; \citealp{liao2019efficient}).

\paragraph{Approaches based on Self Attention}
Graph Neural Networks based on \emph{self-attention} are closely related to MPNNs. 
The main difference to MPNNs is that they use self-attention to compute a \emph{weighted} sum of the incoming messages (based on the relations) for updating the node representations.
This provides a useful mechanism for dynamically adapting the information routing (here a kind of soft variable binding) and thereby infer the desired structure for a fixed set of nodes. 
However, note that this may be computationally inefficient because it still requires computing all possible messages and only affects which of them end up being used in the final summation.
\Citet{wang2018nonlocal} makes use of a kind of (learned) dot-product attention to infer relations between spatial slots.
In this case, the attention coefficients are computed for pairs of nodes while the messages are based only on a single node, which may make it more difficult to implement multiple different relations.
The use of \emph{multiple attention heads}~\citep[\ie as in][]{vaswani2017attention} may help mitigate this issue and has been successfully applied for relational reasoning about objects \citep{zambaldi2019deep,vansteenkiste2019investigating,goyal2019recurrent,santoro2018relational}, citation networks~\citep{velickovic2018graph}, question answering~\citep{dehghani2019universal}, and language modeling~\citep{devlin2018bert,brown2020language}.
Indeed, Transformers themselves may already be viewed as a kind of graph network~\citep{battaglia2018relational}.
Alternatively, multiple different relations could be learned by also conditioning the message on the receiving object representation when using attention \eg as in R-NEM~\citep{vansteenkiste2018relational}.
The idea of using (self-)attention as a mechanism for inferring structure (and dynamic information routing) has also been applied outside the scope of graph neural networks, \eg in pointer networks~\citep{vinyals2015pointer}, energy-based models~\citep{mordatch2019concept}, and capsules~\citep{sabour2017dynamic,kosiorek2019stacked}.

\subsubsection{Neural Computers}
\label{sec:comp:neural_computers}

Neural computers offer an alternative approach to composition by learning to perform reasoning operations sequentially on some appropriate representation of the desired structure.
In this case, the ‘processor’ is typically given by an RNN that interfaces with other components, such as a dedicated memory, via a prescribed set of differentiable operations.
Compared to a GNN, the architecture of a neural processor is more generic and does not directly reflect the desired dependency structure in terms of relations between object representations.
Instead, by considering structure at a representational level, it can more easily be adjusted depending on task or context.
Similarly, by having a \emph{central} processor that is responsible for relational responding (as opposed to a distributed GNN) it is easier to support operations that require \emph{global} information (\eg structure-sensitive operations).
On the other hand, the ability of neural computers to learn more general algorithms comes at the cost of a weaker inductive bias for relational reasoning specifically.
Hence, it is often necessary to incorporate more specialized mechanisms to efficiently learn algorithms for relational responding that generalize in agreement with the desired structure.

The most common type of neural computer consists of an RNN (the processor) that interfaces with an external differentiable memory component.
A dedicated memory component provides an interface for routing information content (now stored separately) to the \emph{variables} that take part in processing (\ie the program executed by the RNN processor).
Indeed, while an RNN can in principle perform any kind of computation using only its hidden state as memory~\citep{siegelmann1991turing}, its dual purpose for representing structure and information processing makes it difficult to learn programs that generalize systematically~\citep{lake2018generalization,csordas2020are}.
Early examples of memory-augmented RNNs~\citep{das1992learning,mozer1993connectionist} use a continuous adaptation of stacks based on the differentiable push and pop operations introduced by \citet{giles1990higher} (\cf \citealp{joulin2015inferring} for an alternative implementation).
Although a stack-based memory has proven useful for learning about the grammatical structure of language\citep[\eg][]{das1992learning}), its utility for more general reasoning tasks is limited by the fact that only the top of the stack is accessible at each step. 

The addressable memory used in the Neural Turing Machine (NTM;  \citealp{graves2014neural}) offers a more powerful alternative, which can be accessed via generic read and write operations (but see memory networks for a read-only version;~\citealp{weston2014memory,sukhbaatar2015endtoend}).
In this case, all memory slots (and thereby all parts of the structure) are simultaneously accessible through an attention mechanism (responsible for variable binding) that supports both content- and location-based addressing. 
Together, these operations have shown to provide a useful inductive bias for learning simple algorithms (\eg copying or sorting) that generalize to longer input sentences (\ie more systematically).
Additional memory addressing operations, \eg based on the order in which memory locations are accessed (DNC;~\citealp{graves2016hybrid}), based on when they were last read~\citep{munkhdalai2017neural}, or based on a key-value addressing scheme~\citep{csordas2019improved} may confer additional generalization capabilities that are especially relevant for relational reasoning.
For example, the DNC has shown capable of learning traversal and shortest path algorithms for general graphs by writing an input sequence of triples (‘from node’, ‘to node’, ‘edge’) to memory, and iteratively traverse this structure using content-based addressing~\citep{graves2016hybrid}.
Moreover, given a family tree consisting of ancestral relations between family members, the DNC can successfully derive relationships between distant members, which demonstrates a form of combinatorial entailment. 

Other memory-based approaches take a step towards GNNs by updating each memory location in parallel~\citep{henaff2016tracking,kaiser2015neural} or incorporate specialized structure for reasoning into the processor, \eg for the purpose of visual question answering using a read-only memory (knowledge base; see \citealp{hudson2018compositional}).
Alternatively, certain (Hebbian) forms of fast weights \citep{schmidhuber1992learning} can be viewed as a type of \emph{internal} associative memory based on previous hidden states~\citep{ba2016using}.
TPR-RNN~\citep{schlag2018learning} extends this idea by equipping a fast-weight memory with specialized matrix operations inspired by Tensor Product Representations (TPR;~\citealp{smolensky1990tensor}), which makes it easier to respond to relational queries.  
In contrast, \citet{reed2015neural} and \citet{kurach2016neural} take a step towards modern computer architectures by, respectively, incorporating a call-stack with an explicit compositional structure or a mechanism for manipulating and dereferencing pointers to a differentiable memory tape.\looseness=-1

\subsection{Learning and Evaluation}
\label{sec:comp:summary}
The problem of composition is about implementing structured models with neural networks that take advantage of the underlying compositionality of object representations.
We have argued that this requires incorporating mechanisms for dynamic variable binding (such as attention), and for dynamically organizing internal information processing for the purpose of relational responding.
Regarding the latter, the choice of a suitable mechanism is less clear, although evidence indicates that a GNN-based approach is promising.

With the right mechanisms in place, it is reasonable to expect that relations, relational frames, and structure inference can all be learned (jointly with segregation and representation) via mostly unsupervised learning~\citep{kemp2008discovery}. 
On the other hand, learning about relations in particular may be challenging, since they can never be observed directly, but always occur in conjunction with concrete objects.
Indeed, young children initially reason primarily based on the perceptual similarity between objects and learn to pay attention to their relational similarity only at a later stage (\ie after undergoing a ``relational shift’’; \citealp{gentner1991language}).
A key enabler for children to acquire progressively more general relations is multi-exemplar training: repeated exposure to the same relation, but in combination with different fillers~\citep{barnes-holmes2004establishing,luciano2007role}.
This idea has been successfully adapted for learning abstract relations using spiking neural networks~\citep{doumas2008theory}, and shares similarities to more recent contrastive learning objectives that require a neural network to infer relations from a dataset of positive and negative pairings~\citep{kipf2019contrastive,hadsell2006dimensionality}.
The ability to interact with the environment may additionally enable an (embodied) agent to autonomously acquire multi-exemplar data for a \emph{particular} relation~\citep{schmidhuber2015learning,haber2018learning}.
An alternative approach to learning composition is to view dynamic structure inference as a meta-learning problem and directly optimize for (systematic) generalization, \eg by minimizing the generalization regret in face of deliberate  non-stationarity~\citep{bengio2019metatransfer}.\looseness=-1

The ultimate goal of composition is to facilitate more systematic generalization and several methods have been proposed that measure different aspects of this ability.
A prototypical approach is to evaluate a trained system on a set of held-out combinations of parts (objects) as an approximate measure of systematicity~\citep{santoro2018measuring, lake2018generalization,hupkes2019compositionality}.
A similar strategy can also be used to assess the capacity for interpolation or extrapolation, \ie by varying the number of parts or range of values.
Additionally, \citet{hupkes2019compositionality} propose to measure (systematic) ``overgeneralization errors’’ that are indicative of a bias towards a particular pattern of generalization.\looseness=-1

}

{
\section{Insights from Related Disciplines}
\label{ch:rel}

Object perception and the symbolic nature of human cognition have been studied from various angles in Neuroscience, Psychology, Linguistics, and Philosophy.
These complementary perspectives provide valuable inspiration for addressing the binding problem and we have frequently drawn upon their insights throughout this survey.
While an exhaustive overview is outside the scope of this survey, we provide a brief discussion of the areas that were most influential to the development of the conceptual framework presented here.
These fields have a lot more to offer and we encourage the reader to further explore this literature, for example by using the pointers and connections provided here as entry-points.

\begin{SCfigure}[][t]
    \centering
    \includegraphics[width=0.7\textwidth]{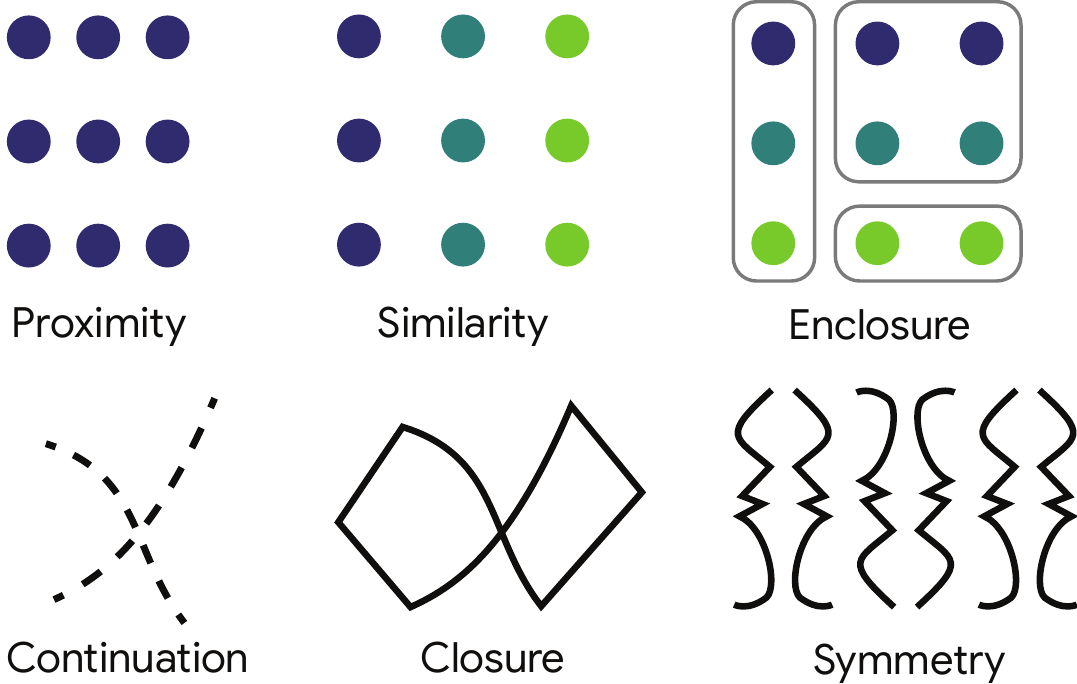}
    \caption{Illustration of several Gestalt Laws of visual perception. 
    Note how the different cues influence which elements are perceived as belonging together.}
    \label{fig:rel:gestalt_laws}
\end{SCfigure}

\subsection{Gestalt}
\label{sec:rel:gestalt}

Gestalt Psychology describes many aspects of the subjective experience of perceptual organization (see \citealp{wagemans2012century, wagemans2012centurya} for an overview).
It is based on the observation that the perception of `wholes’ (or Gestalten\footnote{
\emph{“Gestalten”} is plural of the German word \emph{“Gestalt”} meaning “form” or “shape”.}) can not be adequately described as a bottom-up agglomeration of more primitive percepts, but rather emerges in its entirety at once. 
Similarly, the perception of a Gestalt can fill in missing information, be invariant to transformations, and alternate discretely between multiple stable interpretations (see \cref{fig:rel:gestalt_properties}).
This holistic (as opposed to analytic; see \cref{sec:rel:fit}) view of perception, was later summarized by Kurt Koffka as: “The whole \emph{is other} than the sum of its parts”~\citep{koffka1935principles}\footnote{
Frequently misquoted as “The whole is \emph{greater} than the sum of its parts”.}.
The concept of a Gestalt closely resembles our notion of objects and Gestalt Psychology was arguably the first systematic investigation of human object perception (following the work by \citealp{wertheimer1912experimentelle}).

The best-known results of Gestalt research are their principles of perceptual grouping (also known as Gestalt Laws; see \cref{fig:rel:gestalt_laws} for an overview).
They describe which stimulus-cues influence the perceived grouping of a set of discrete elements~\citep{wertheimer1923untersuchungen,wagemans2012century}.
They include among others: the law of proximity (closeby pieces tend to be grouped), the law of similarity (similar pieces tend to be grouped), the law of closure (preference for closed contours), the law of symmetry (preference for symmetric objects) and the law of common fate (what moves together groups together).
Several other Gestalt laws have been found over the years~\citep{alais1998visual,palmer1992common,palmer1994rethinking}, including for other sensory modalities, such as audio~\citep{bregman1994auditory} and tactile~\citep{gallace2011what}.
Note that the laws of proximity and common fate can be seen as special cases of the law of similarity (with position and movement respectively being the compared attributes).
In fact, it has been argued that the Gestalt Laws are all special cases of a single information-theoretic grouping principle~\citep{hatfield1985status}.
Here the idea is that a `good’ Gestalt is one with a lot of internal redundancy~\citep{attneave1971multistability}, and thus that the likelihood of a particular grouping is inversely proportional to the amount of information required to describe the Gestalt~\citep{hochberg1953quantitative}\footnote{
There is disagreement about how to quantify information and the issue of simplicity versus likelihood has been debated extensively, though they might turn out to be identical~\citep{chater1996reconciling}}.

\begin{SCfigure}[40]
    \includegraphics[width=0.45\textwidth]{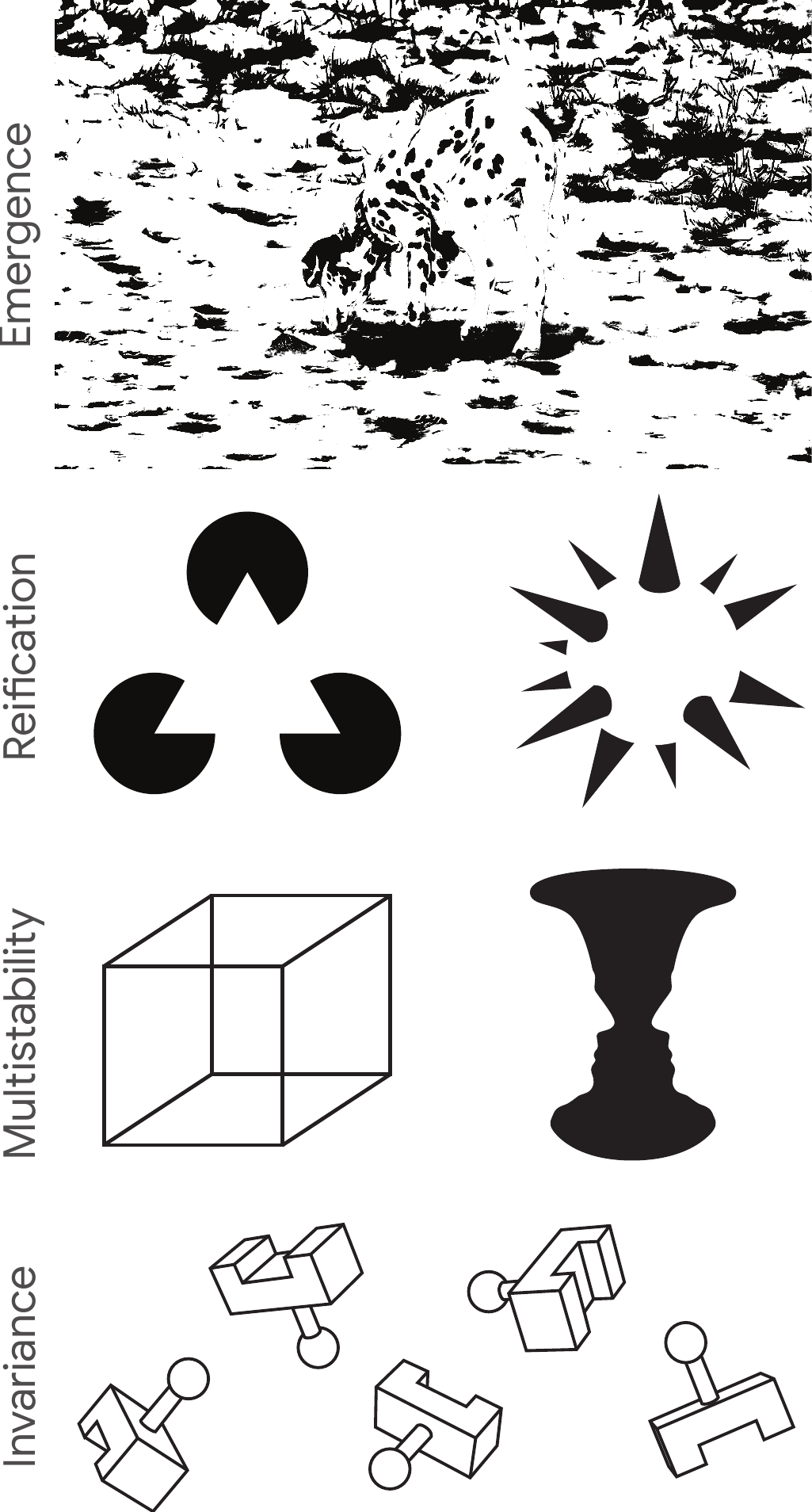}
    \caption{
    \textbf{Emergence:} 
        At first encounter this image (a reproduction of the classic image from \citealt{gregory1970intelligent}) is perceived as an unstructured collection of black patches on white background.
        At some point perception shifts and suddenly reveals the image of a Dalmatian dog sniffing the ground. 
        Perception of the whole arises at once, and not through hierarchically assembling of parts, such as legs, ears, \etc. \\[1em]
        \textbf{Reification:} Perception of a Gestalt carries information about its parts and leads to top-down ``filling-in'' of missing information.
        This is often demonstrated with the illusory contours of the \emph{Kanizsa triangle} (left).\\[2em]
        \textbf{Multistability:} Many scenes are ambiguous and afford multiple stable groupings. In such cases perception alternates periodically between different interpretations.\\[2em]
        \textbf{Invariance:} Objects are recognized based on their overall shape invariant of: rotation, shift, scale, illumination, and many other factors.
    }
    \label{fig:rel:gestalt_properties}
\end{SCfigure}

For our purposes, the existence of these general principles and their prevalence in multiple sensory domains is very interesting.
It makes plausible the idea of a general segregation mechanism (\eg based on modularity) that can generalize to novel objects and can help to steer the search for corresponding inductive biases.
However, note that Gestalt Psychology has been criticized for its emphasis on subjective experience and the lack of successful physiological or mechanistic predictions (\eg \citealp{ohlsson1984restructuring,treisman1980featureintegration}; but see \citealp{jakel2016overview}).
Feature Integration Theory arose as a countermovement to provide an alternative, more mechanistic, account of the grouping process.

\subsection{Feature Integration Theory}
\label{sec:rel:fit}

Feature Integration Theory (FIT) provides a model of human visual attention for perceiving objects (see \citealp{wolfe2020forty} for an overview).
It is based on the idea that conscious object perception (\ie as we experience it) is preceded by subconscious (mostly) bottom-up processing of visual information.
FIT is motivated by a number of empirical findings, such as the different speeds at which humans are able to locate a visual target among a set of distractors (visual search).
In this case, search is fast (subconscious and in parallel) if the target can be identified by a single characteristic feature (\eg a particular orientation), which essentially causes it to \emph{pop-out} (\eg top panel in \cref{fig:rel:fit}a).
In contrast, when the target is characterized by a \emph{conjunction} of features, search becomes slow and requires serial attention (\eg bottom panel in \cref{fig:rel:fit}a).
Another important empirical finding occurs when attention is overloaded (or directed elsewhere), which sometimes causes humans to perceive \emph{illusory conjunctions}: illusory objects that are the result of wrongly combining features from other objects (\citealp{treisman1980featureintegration}; \cref{fig:rel:fit}c).

Feature Integration Theory distinguishes two stages of processing (see \cref{fig:rel:fit}b).
First, a \emph{pre-attentive stage} that registers features across the visual field (\eg shape, color, size, \etc) automatically in parallel, and represents them in independent feature maps (‘free-floating’).
Then, at the \emph{feature integration stage}, a `spotlight of attention’ is used to bind the features in these separate maps to form feature conjunctions in the form of objects~\citep{kahneman1992reviewing}.
While initially objects are linked to specific locations as attention is focused on them, they may later be consolidated to form a more location invariant representation~\citep{treisman2006location}.
Since its initial conception~\citep{treisman1977focused,treisman1980featureintegration}, FIT has been refined and extended in various ways to account for new insights about human perception.
There is now substantial evidence that the features of objects outside the focus of attention are more structured than initially assumed.
For example, \citet{humphrey1998probing} find that orientation and color are already represented jointly in the absence of attention (see also \cite{vul2006contingent}).
Similarly, \citet{vul2019structure} find evidence for pre-attentive binding of color to parts based on the hierarchical (and geometric) structure of objects.

\begin{figure}
    \centering
    \includegraphics[width=0.95\textwidth]{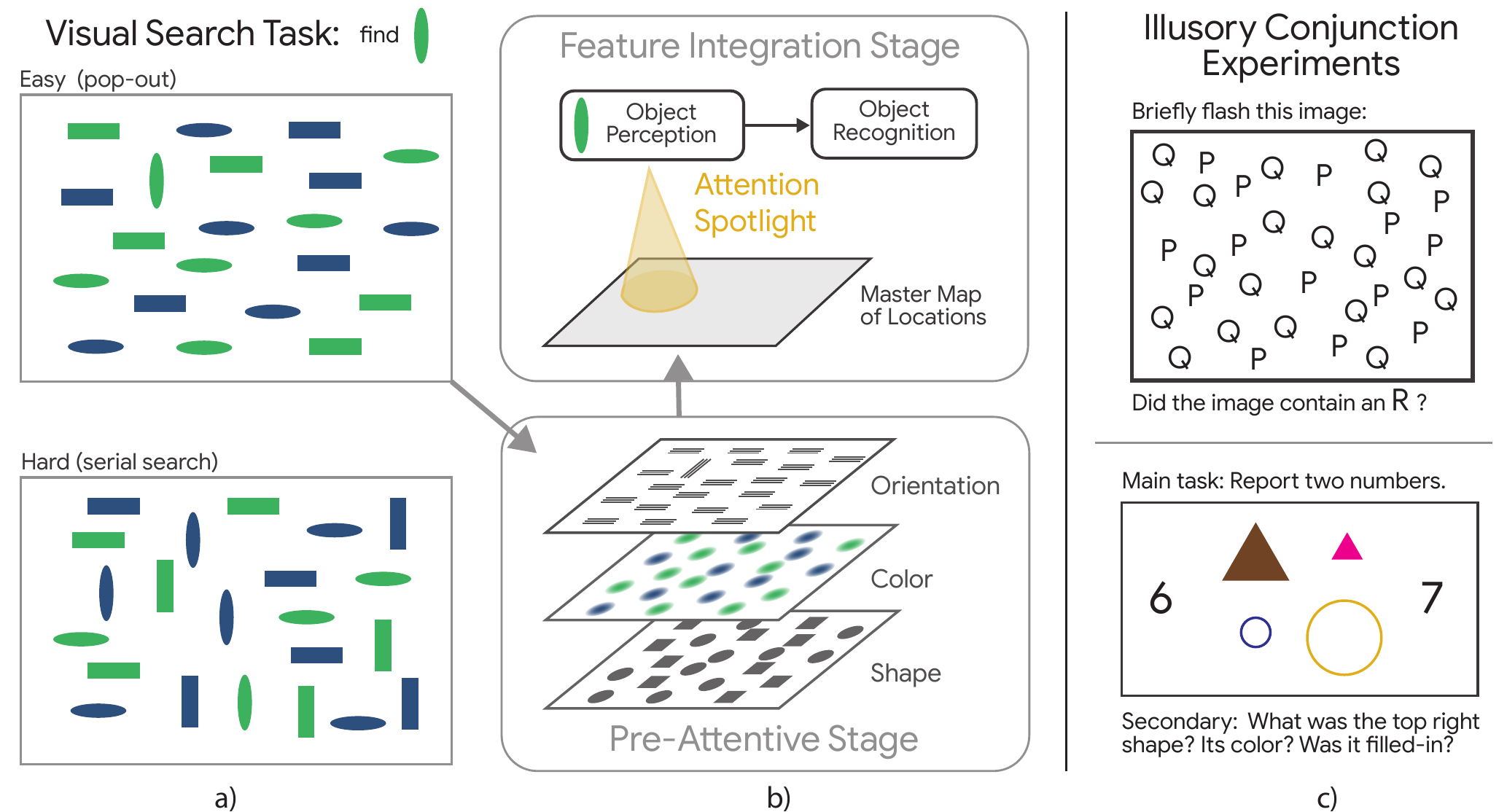}
    \caption{
        \textbf{Left:} Two examples of visual search tasks, an easy one where the target ``pops-out'' (top) and a hard one that requires serial search. 
        \textbf{Middle:} Diagram of processing operations involved in the perception of objects according to FIT.
        \textbf{Right:} Two example tasks that have been used to demonstrate \emph{illusory conjunctions}. Note, that the effect cannot be reproduced in print because it relies on showing the images very briefly.
    }
    \label{fig:rel:fit}
\end{figure}

FIT has been a highly influential model of human visual attention and could serve as further inspiration for attention-based segregation. 
While FIT and Gestalt Psychology offer seemingly competing views of human perception, it has also been argued that these analytic and holistic views in fact complement each other~\citep{prinzmetal1995visual}.
However, in either case, it is unclear how certain aspects of FIT should be implemented, such as top-down feedback to guide attention, especially in the context of non-visual domains~\citep{spence2020multisensory}.

\subsection{The Binding Problem in Neuroscience}
\label{sec:rel:binding_problem}

We have adapted the term binding problem from neuroscience, where it refers to a limitation of our understanding regarding information processing in the brain.
In particular, its highly distributed nature raises the question of ``[\dots] how the computations occurring simultaneously in spatially segregated processing areas are coordinated and bound together to give rise to coherent percepts and actions’’ --- \citet{singer2007binding}.
For example, how is it that we typically do not wrongly mix the properties belonging to different objects, \ie experience illusory conjunctions?
The binding problem in neuroscience is thus concerned with understanding the mechanism(s) by which the brain addresses these challenges.

Several mechanisms have been proposed that range from static binding using conjunction cells~\citep{ghose1999specialized} to dynamic information routing through dedicated circuitry~\citep{olshausen1993neurobiological,zylberberg2010brain} or attention using common location tags~\citep{reynolds1999role,robertson2005attention}.
A particularly promising hypothesis is the \emph{temporal correlation hypothesis}, which holds that temporal synchrony of firing patterns is the mechanism responsible for binding~\citep{milner1974model, vondermalsburg1981correlation}.
In this case, neurons whose activation encodes features of one object (\eg color and shape) are expected to fire in synchrony (oscillating phase-locked), while neurons encoding features belonging to different objects would be out of phase with each other (see also \cref{sec:rep:augmentation}).
Other neurons are naturally capable of responding to this form of grouping since neuronal firing and synaptic learning (STDP; \citealp{caporale2008spike}) are both sensitive to the relative timing of incoming activations (pre-synaptic spikes).
Moreover, there is diverse experimental data in support of this interpretation relating synchronized oscillatory behavior of individual neurons to perceptual grouping~\citep{usher1998visual,tallon-baudry1999oscillatory}, attention~\citep{niebur2002synchrony}, and sensory-motor integration~(\citealp{pesceibarra2017synchronization,engel2010betaband}; see also \citealp{uhlhaas2009neural} for an overview).

In general, the role of synchrony in neuronal binding is still controversial.
For example, it has been debated whether synchrony is necessary~\citep{merker2013cortical, riesenhuber1999are}, fast enough~\citep{ray2015gamma, palmigiano2017flexible}, and is capable of providing sufficient (temporal) resolution\footnote{
In this case, the temporal accuracy of synchronization directly relates to the capacity of working memory~\citep{wilhelm2013what}.
The more objects need to be represented simultaneously, the more difficult it is to prevent cross-talk from corrupting and destabilizing individual representation, and such gradual decay has indeed been observed in~\cite{alvarez2007how}.
} for separating multiple different objects.
Likely, the brain does not rely on a single mechanism for addressing the binding problem but on a combination of several. 
In either case, it is clear that temporal synchronization plays an important role in neural information processing, and perhaps one that is still unaddressed in current artificial neural networks.

\subsection{Relational Frame Theory}
\label{sec:rel:rft}

\begin{SCfigure}[30][t]
    \hspace{1.5em}\includegraphics[width=0.21\textwidth]{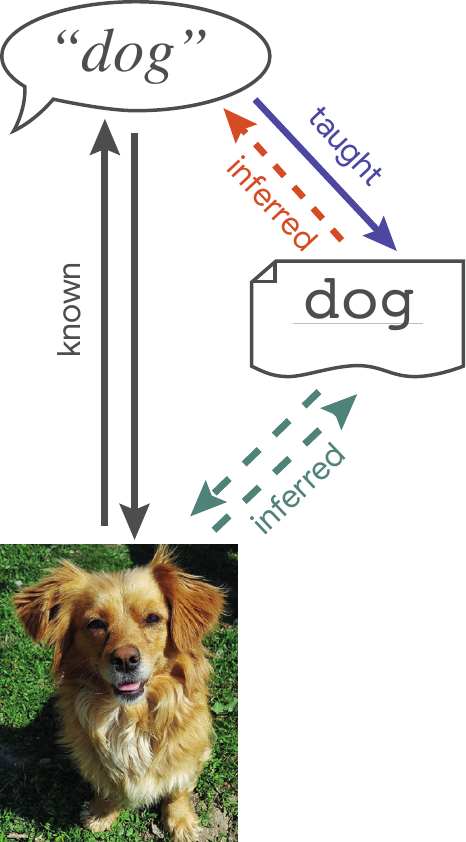}\hspace{1.5em}
    \caption{
        In an early experiment, \citet{sidman1971reading} examined a boy that could match spoken words to pictures and to name pictures (\textcolor{darkgray}{gray}), but was unable to read.
        After being taught to match spoken words to written words (\textcolor{gs_purple}{blue}), he was then also able to read the written words aloud (\textcolor{gs_red}{red}), and to match them to pictures (\textcolor{gs_green}{green}).
        In this case, the dotted arrows represent relations that were never explicitly taught, and which were \emph{derived} based on reflexivity (\textcolor{gs_red}{red}) and transitivity (\textcolor{gs_green}{green}) of the underlying equivalence relation \citep{sidman1989functional}.
        Later, it was found that such derived relationships play an important role in systematically altering human behavior in response to feedback from the environment.
    }
    \label{fig:rel:rft:example}
\end{SCfigure}

Relational Frame Theory (RFT;~\citealp{hayes2001relational,hughes2016relational}) is a theory of behavioral psychology about \emph{relating} (\ie responding to one event in terms of another) and offers interesting insights about composing and systematic generalization in humans.
RFT was originally conceived to explain ``stimulus equivalence’’~\citep{sidman1971reading}:
The emergent behavior to respond to events and objects through a derived ``sameness’’ relation that has not been explicitly taught or reinforced. 
For example, when taught a correspondence between spoken words and pictures, and between spoken words and written words, children were able to match written words and pictures (see \cref{fig:rel:rft:example}).
In a similar experiment, \cite{dymond1995transformation} showed that subjects were able to use such derived equivalence relations to ``correctly’’ respond to stimuli for which no explicit feedback was provided.

RFT is based in behaviorism, focusing on observable behavioral responses that can be altered through reinforcement or punishment (learned operants).
It argues that relational responding \footnote{
RFT distinguishes between two types of relational responding: Non-Arbitrarily Applicable Relational Responding (NAARR), which is only concerned with relations among physical attributes (\eg choosing the \emph{larger} among multiple objects), and the more general Arbitrarily Applicable Relational Responding (AARR) that allows for arbitrary relations between stimuli (or events).
While NAARR is also encountered in animals, AARR has thus far only been observed in humans. 
} is a learned operant behavior, which can be acquired through repeated exposure to tasks that require responding to a particular relation (to receive positive feedback) but that varies across stimuli and contexts.
Relational responding can be subdivided into different \emph{relational frames} (see also \cref{sec:comp:relational_frames}), which each focus on a particular kind of relationship and differ in terms of three key properties: \emph{mutual entailment}, \emph{combinatorial entailment}, and \emph{transformation of stimulus functions}.
For example, the stimulus equivalence that was observed in \cref{fig:rel:rft:example} corresponds to a particular relational frame with symmetry as mutual entailment and transitivity as combinatorial entailment.
In this case, `transformation of stimulus function' implies that when the reward associated with an object or event changes, this also alters the expected reward of other related events or objects in the same manner.
Other examples include the relational frames of `opposition (\eg opposite to) , `comparison' (\eg larger than), `hierarchy' (\eg part of) , `temporal order’ (\eg after), or `condition' (\eg if then).
It is easy to see how a vast number of possible relational structures can be constructed in this way, of which only very few are relevant in any given situation.
RFT argues that people use (bottom-up) \emph{contextual cues} from the environment to infer which relations when to apply.

Given the immediate relevance of RFT to systematic generalization and composition, it is surprisingly absent from the machine learning literature.
This is likely in part due to the relative unpopularity of behaviorism compared to cognitive psychology.
However, another reason may be due to the controversy that surrounds certain aspects of RFT, such as the clarity of the involved concepts and its novelty with regards to previous accounts of stimulus equivalence~\citep{gross2009relational}.
Nonetheless, we find that RFT offers a useful conceptual framework for the problem of composition, and indeed it has helped shape our understanding of relational reasoning. 
Going forward, we would like to emphasize the value of RFT as a source of experimental designs to isolate and evaluate relational reasoning capabilities in neural networks.

\subsection{Compositionality in Linguistics}
\label{sec:rel:compositionality}

Like many others in the field, we have used the term compositionality without giving a proper definition.
Related terms such as systematicity, systematic generalization, and combinatorial generalization, unfortunately, do not provide a good alternative either.
A good starting point for a definition may therefore be the so-called \emph{principle of compositionality} from the field of linguistics:
\begin{quote}
	“The meaning of a complex expression is determined by its structure and the meanings of its constituents.” --- \citet{szabo2017compositionality}
\end{quote}
Apart from its intuitive appeal, the main reason for its widespread adoption is the lack of a convincing alternative.
However, there remains considerable disagreement about the exact phrasing and many interpretations of the principle exist~\citep{szabo2017compositionality}.

There are three main arguments for this notion of compositionality, namely \emph{productivity}, \emph{systematicity}, and \emph{efficiency} of language.
Productivity refers to the capacity of language to “make infinite use of finite means”~\citep{vonhumboldt1999humboldt}, \ie the ability to form and understand a theoretically unbounded number of entirely novel sentences given only limited vocabulary and training.
Systematicity is the observation that “the ability to produce/understand some sentences is intrinsically connected to the ability to produce/understand certain others”~\citep{fodor1988connectionism}.
For example, anyone who understands “brown dog” and “black cat” also understands “brown cat”.
Finally, the fact that we are able to communicate in real-time, puts clear bounds on the computational complexity of interpreting spoken language~\citep{szabo2017compositionality}.
The principle of compositionality is thus an inference to the best explanation because it is difficult to imagine language being productive, systematic, and computationally efficient without its semantics being somehow compositional in the above sense.

Critique of the principle of compositionality, interestingly, ranges from it being too broad to it being too narrow.
On the one hand, \citet{zadrozny1994compositional} demonstrates how a function can be constructed that maps arbitrary meaning to any expression without violating compositionality.
This suggests that the principle is formally vacuous unless the class of admissible functions is somehow restricted to exclude such a construction.
On the other extreme, many have found violations of the principle in everyday language.
Indeed, counterexamples such as ambiguities (“We saw her duck.”), references (“this dog”), and irony (“objectively the best example”) require context and thus contradict the principle.
Similarly, idioms (“break the ice”) provide examples of obvious exceptions where the meaning differs substantially from a naive composition of the parts. 
However, few consider these problems severe enough to abandon the principle of compositionality entirely, and indeed most linguists have come to accept it as a guiding principle for developing syntactic and semantic theories.
Though it was originally conceived for language, many believe that the principle of compositionality applies equally (or even more so) to mental representations~\citep{butler1995content,fodor1975language}.
A similar belief also underlies the interest in compositionality for understanding and encouraging productivity, systematicity, and efficient inference in neural networks~\citep{santoro2018measuring,hupkes2019compositionality,andreas2019measuring}.
}

{

\section{Discussion}
\label{sec:discussion}

The ultimate motivation of this work is to address the shortcomings of neural networks at human-level generalization.
To this end, we have developed a conceptual framework centered around compositionality and the binding problem.
Our analysis identifies the binding problem as the primary cause for these shortcomings, and thereby paves the way for a single unified solution.
It rests on several (implicit) assumptions regarding the nature and importance of objects and the learning capabilities of neural networks.
In the following, we explicate several of these assumptions and use them to contrast with other conceptual frameworks aimed at addressing (certain aspects of) human-level generalization.

One of the main assumptions behind our work is that objects are key to compositionality and that the latter plays a fundamental role in generalizing more systematically.
This perspective has a long history in connectionism that goes back to at least~\citet{fodor1988connectionism, marcus2003algebraic} and has been repeatedly emphasized~\citep[\eg][]{smolensky1990tensor,bader2005dimensions}, especially in recent years~\citep[\eg][]{lake2017building,battaglia2018relational,hamrick2019analogues, garnelo2019reconciling}.
However, our perspective stands out in that we focus on integrating symbolic reasoning and sensory grounding, which requires adopting a very broad notion of objects that spans all levels of abstraction.
Importantly, we assume that objects at any level of abstraction are essentially the result of decomposing a given problem into modular building blocks, and thus share the same underlying computational mechanisms.
It is our view that this broad notion of objects is necessary to accommodate the generality of human reasoning from concrete and physical to abstract and metaphorical.

Throughout this paper, we have assumed that learning objects in an unsupervised way is both feasible, and can be integrated directly into neural networks.
Further, we have argued that unsupervised learning is, in fact, indispensable, due to the required scope and flexibility of objects, which renders adequate supervision or engineering infeasible.
However, as we have seen (and discuss further below), evidence indicates that object representations are unlikely to emerge naturally simply by scaling current neural networks in terms of model size or by providing additional data.
Here we have proposed to address this problem by incorporating a small set of inductive biases to enable neural networks to process information more symbolically, while also preserving the crucial benefits of end-to-end learning \citep{sutton2019bitter}.

Closely related to the mental framework proposed here is that of \citet{lake2017building}, which is similarly concerned with addressing human-level generalization.
They too emphasize the importance of (physical) objects, compositionality, and dynamic model building, although in their view these are only three instances of so-called `core ingredients’ necessary for realizing human intelligence.
Other ingredients include an intuitive understanding of psychology as a form of ``start-up software’’, learning to learn, causality, and ingredients focusing on the speed of human comprehension.
Hence, \citet{lake2017building} advocate the use of specialized inductive biases inspired by cognitive psychology, and using neural networks as a \emph{means} for implementing fast inference within the context of larger structured models. 
In contrast, we argue that it is more fruitful to enable neural networks to directly implement structured models.
This enables us to tackle a single shared underlying problem (the problem of dynamic binding) and, as much as possible, let learning account for the remaining, domain-specific, aspects of human cognition (\eg psychology, physics, causality).
Note that we do not wish to argue against incorporating specialized inductive biases (which may still be beneficial), but rather advocate that learning should take priority whenever possible. 
Compared to \citet{lake2017building} our focus on integrating high-level reasoning with low-level perception in neural networks puts a lot more emphasis on symbol grounding and the associated problem of segregation. 
This is also reflected by our emphasis on end-to-end learning, whereas \citet{lake2017building} appear to argue for separating neural and symbolic information content, somewhat akin to hybrid approaches~\citep{bader2005dimensions}.  

Our framework also relates to several other areas of machine learning research that aim towards human-level generalization.
However, they center around composition and have mostly neglected the problem of segregation (and representation).
For example, the field of causality is concerned with inferring and reasoning about structural causal models, which offer a particular kind of compositionality that is assumed to be essential to human-level generalization~\citep{pearl2009causality,peters2017elements}.
Using our terminology, structural causal models can be viewed as a specific set of relational frames composed of `independent causal mechanisms’ that define a structure, which can be used to systematically reason about novel situations (\eg for interventions or counterfactuals).
As was recently noted by \citet{scholkopf2019causality}, traditional work in causality assumes given knowledge about the associated causal variables (\eg objects), and the problem of discovering them (\ie segregation) has mostly been neglected. 
In a similar vein, recent work on graph neural networks seeks to achieve systematic generalization by focusing on relations between \emph{given} entities~\citep{battaglia2018relational}. 
Alternatively, \citet{bengio2019consciousness} argues for the importance of a low-dimensional `conscious state’ (working memory) composed of largely independent units of abstraction that can be selected via attention (perhaps reminiscient of \citealp{schmidhuber1992learninga}).
He relates the unconscious elements from which the conscious state is constructed to a more symbolic knowledge representation, and emphasizes their importance for systematic generalization.
However, here too, it remains unclear how such elements should be obtained and represented in neural networks.

Finally, we acknowledge the promising results that recent large-scale language models have produced in terms of generalization and their (acquired) ability for few-shot learning~\citep{radford2019language,brown2020language}.
They are evidence for the possibility that human-level generalization may be achieved by scaling existing approaches using orders of magnitude more data and network parameters.
However, we remain pessimistic as to whether similar results can be obtained on less structured domains, such as when learning from raw perceptual data.
As we have argued throughout this work, the fundamental lack of a suitable mechanism for dynamic information binding precludes the emergence of the modular building blocks needed for acquiring a compositional understanding of the world.
}
{

\section{Conclusion}
\label{sec:conclusion}

Humans understand the world in terms of abstract entities, like objects, whose underlying compositionality allows us to generalize far beyond our direct experiences.
At present, neural networks are unable to generalize in the same way.
In this paper, we have argued that this limitation is largely due to the binding problem, which impairs the ability of neural networks to effectively incorporate symbol-like \emph{object representations}.
To address this issue, we have proposed a functional division of the binding problem that focuses on three different aspects:
The ability to separately represent multiple object representations in a common format, without interference between them (representation problem);
The process of forming grounded object representations that are modular from raw unstructured inputs (segregation problem);
And finally, the capacity to dynamically relate and compose these object representations to build structured models for inference, prediction, and behavior (composition problem).
Based on this division, we have offered a conceptual framework for addressing the lack of symbolic reasoning capabilities in neural networks that is believed to be the root cause for their lack of systematic generalization.
Indeed, the importance of symbolic reasoning has been emphasized before~\citep{fodor1988connectionism} and served as a starting point for several related perspectives~\citep{marcus2003algebraic,lake2017building}.
Here we have provided a more in-depth analysis of the challenges, requirements, and corresponding inductive biases required for symbol manipulation to emerge naturally in neural networks.

Based on our discussion, we wish to highlight several important open problems for future research in three different areas.

First is the process of segregation, which is of foundational importance and requires a proper treatment of the dynamic and hierarchical nature of objects.
In particular, we believe that the ability to segregate must therefore largely be learned in an unsupervised fashion, which is a major open problem that is often overlooked in the current literature.
For a new situation, the most useful decomposition in terms of objects (and the associated level of abstraction) depends not only on the task, but also on the abstractions, relations, and general problem-solving capabilities available to the entire system.
Therefore, another open problem is to integrate segregation, representation, and composition into a single system in a way that resolves these dependencies (through top-down feedback).
Existing attempts fail to accommodate these interactions, \eg because they rely on pre-trained vision modules~\citep{mao2019neurosymbolic} or overly specialized domain-specific components~\citep{deavilabelbute-peres2018endtoend}.
Addressing these open problems may pave the way for an integrated system that can learn to dynamically construct structured models for prediction, inference, and behavior in a way that generalizes similarly to humans.

Secondly, to facilitate progress on the binding problem, we require corresponding benchmarks and metrics that allow for meaningful comparisons.
Current benchmarks fall short in the sense that they do not bridge the gap between simplistic `toy’ datasets and the complexity of real-world sensory information, or lack the appropriate meta-data required to support evaluation (such as object-level annotations).
The latter is particularly important since standard approaches to measuring properties such as systematic generalization or disentanglement are supervised and require information about `ground truth’ objects or factors.
However, this reliance on ground truth data seems problematic in real-world settings more generally, \ie due to the task- and context-dependent nature of objects and the amount of manual labor involved.
This should motivate research on alternative `unsupervised metrics’ for these purposes, \eg analogous to the FID score for the perceptual quality of images~\citep{heusel2017gans}. 
The design of benchmarks and metrics is hindered by a lack of agreed-upon definitions for behaviors like systematic generalization, combinatorial generalization, or compositionality.
Going forward, it is therefore critical to develop a shared vocabulary of well-defined and measurable generalization patterns that can be explicitly characterized in terms of the type and amount of available information.
Recent attempts at quantifying systematic generalization that distinguish between interpolation and extrapolation~\citep{santoro2018measuring} or the categorization developed by \citet{hupkes2019compositionality} provide a promising step in this direction.

Finally, we wish to highlight several other interesting research directions that are also important for human-level generalization but go beyond the scope of this survey.
Concerning the binding problem, we focused primarily on encoding information about objects in working memory, although similar problems arise in the context of long-term memory.
We speculate that several of the same insights can be applied here, \eg memory recall as a type of segregation, or the need for a separation between relations and objects. 
However, for other challenges, such as the problem of representing information in a scalable way (despite a constantly evolving representational format), the connection is less clear.
Another interesting direction is concerned with the arising and grounding of more abstract concepts like “mammals”, “capitalism” or “a transaction”.
Although abstract objects may be more difficult to obtain, since they are further removed from sensory reality, it is precisely because of this gap that they are capable of participating in a wider range of situations. 
Indeed, this research direction is highly relevant to the broader problem of grounding language, which is concerned with abstract concepts in their most general form.
In this context, it is interesting to note that most (if not all) abstract concepts seem to be grounded in basic physical metaphors~\citep{lakoff2008metaphors}.
Finally, a comprehensive treatment of causal reasoning likely goes beyond composition and should include an explicit treatment of interventions and the ability to reason about hypothetical or unobserved scenarios (counterfactuals). 
This is especially relevant due to the connection between systematic generalization and the increased robustness when considering so-called independent causal mechanisms~\citep{peters2017elements}.
If a suitable causal relational frame can be learned, then this may allow the problem of planning to be phrased as connecting a current state and an imagined goal state, by means of combinatorial entailment.

We hope that this survey may serve as an inspiration and a guide for future work towards achieving human-level generalization in neural networks and that it may spark fruitful discussions that bridge the gap between related fields.

}

\section*{Acknowledgements}

We wish to thank Pina Merkert and Mike Mozer in particular, for their constructive feedback and support.
We also wish to thank Sungjin Anh, Boyan Beronov, Paul Bertner, Matt Botvinick, Alexey Dosovitskiy, Sylvain Gelly, Leslie Kaelbling, Thomas Kipf, Alexander Lerchner, Paulo Rauber, Aleksandar Stani\'c, and Harri Valpola. 
Finally, we wish to thank many other colleagues and friends for useful discussions about binding throughout the last years.
This research was supported by Swiss National Science (SNF) grant 200021\_165675/1 (successor project: no: 200021\_192356) and EU project “INPUT” (H2020-ICT-2015 grant no. 687795).

{\small
\bibliography{BindingSurvey}
}

\end{document}